\documentclass[10pt, onecolumn]{article} 
\usepackage{latex8}
\usepackage{times}
\usepackage{graphicx}
\usepackage{subfigure}
\usepackage{amsmath}
\usepackage{amssymb}
\usepackage{url}

\begin{document}

\title{A Reconstruction Error Formulation for Semi-Supervised Multi-task and Multi-view Learning}

\author{
Buyue Qian, Xiang Wang, and Ian Davidson \\
Department of Computer Science\\
University of California, Davis\\
Davis, CA 95616 \\
\texttt{\{byqian,xiang,indavidson\}@ucdavis.edu}
}

\maketitle
\thispagestyle{empty}

\begin{abstract}
A significant challenge to make learning techniques more suitable for general purpose use is to move beyond i) complete supervision, ii) low dimensional data, iii) a single task and single view per instance. Solving these challenges allows working with ``Big Data" problems that are typically high dimensional with multiple (but possibly incomplete) labelings and views. While other work has addressed each of these problems separately, in this paper we show how to address them together, namely \emph{semi-supervised dimension reduction for multi-task and multi-view learning} (SSDR-MML), which performs optimization for dimension reduction and label inference in semi-supervised setting. The proposed framework is designed to handle both multi-task and multi-view learning settings, and can be easily adapted to many useful applications. Information obtained from all tasks and views is combined via reconstruction errors in a linear fashion that can be efficiently solved using an alternating optimization scheme. Our formulation has a number of advantages. We explicitly model the information combining mechanism as a data structure (a weight/nearest-neighbor matrix) which allows investigating fundamental questions in multi-task and multi-view learning. We address one such question by presenting a general measure to quantify the success of simultaneous learning of multiple tasks or from multiple views. We show that our SSDR-MML approach can outperform many state-of-the-art baseline methods and demonstrate the effectiveness of connecting dimension reduction and learning. 
\end{abstract}

\noindent\textbf{Keywords:} Semi-Supervised Learning; Multi-task Learning; Multi-view Learning; Dimension Reduction; Reconstruction Error.

\section{Introduction} \label{sec:intro}

Three core challenges to making data analysis better suited to real world problems is learning from: i) partially labeled data, ii) high dimensional data and iii) multi-label and multi-view data. Where as existing work often tackles each of these problems separately giving rise to the fields of semi-supervised learning, dimension reduction and multi-task or multi-view learning respectively, we propose and show the benefits of addressing the three challenges simultaneously. However, this requires a problem formulation that is efficiently solvable and easily interpretable. We propose such a framework which we refer to as semi-supervised dimension reduction for multi-task and multi-view learning (SSDR-MML). 

\begin{figure*}[!ht]
\begin{center}
\includegraphics*[width=0.3\textwidth]{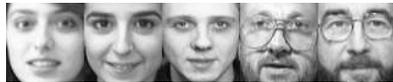}
\caption{Sample face images} \label{fig:faces}
\end{center}
\end{figure*}

\begin{figure*}[!ht]
\begin{center}
\subfigure[PCA]{\label{fig:toy_1}
\includegraphics*[width=0.32\textwidth]{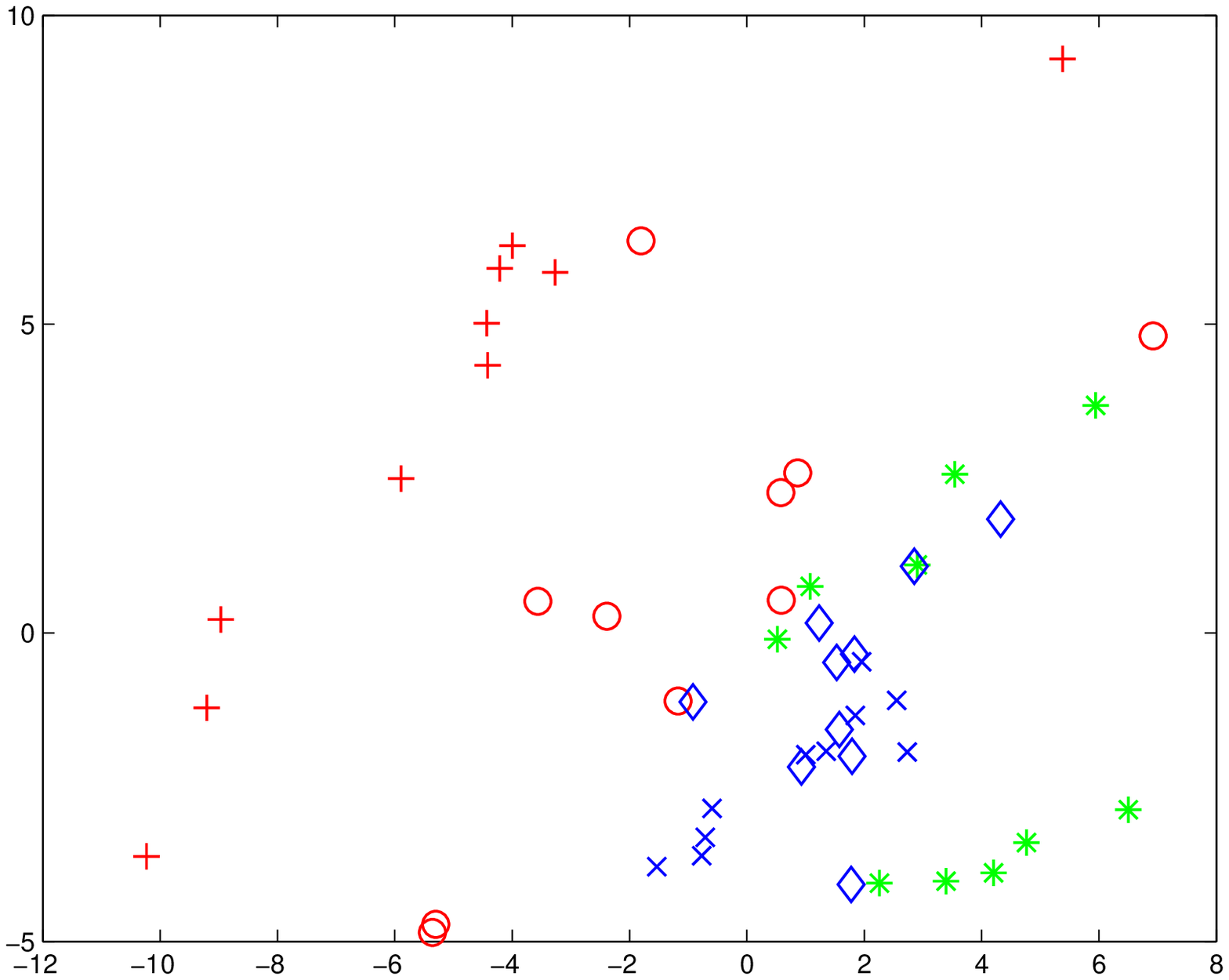}}
\subfigure[PCA+LDA]{\label{fig:toy_2}
\includegraphics*[width=0.32\textwidth]{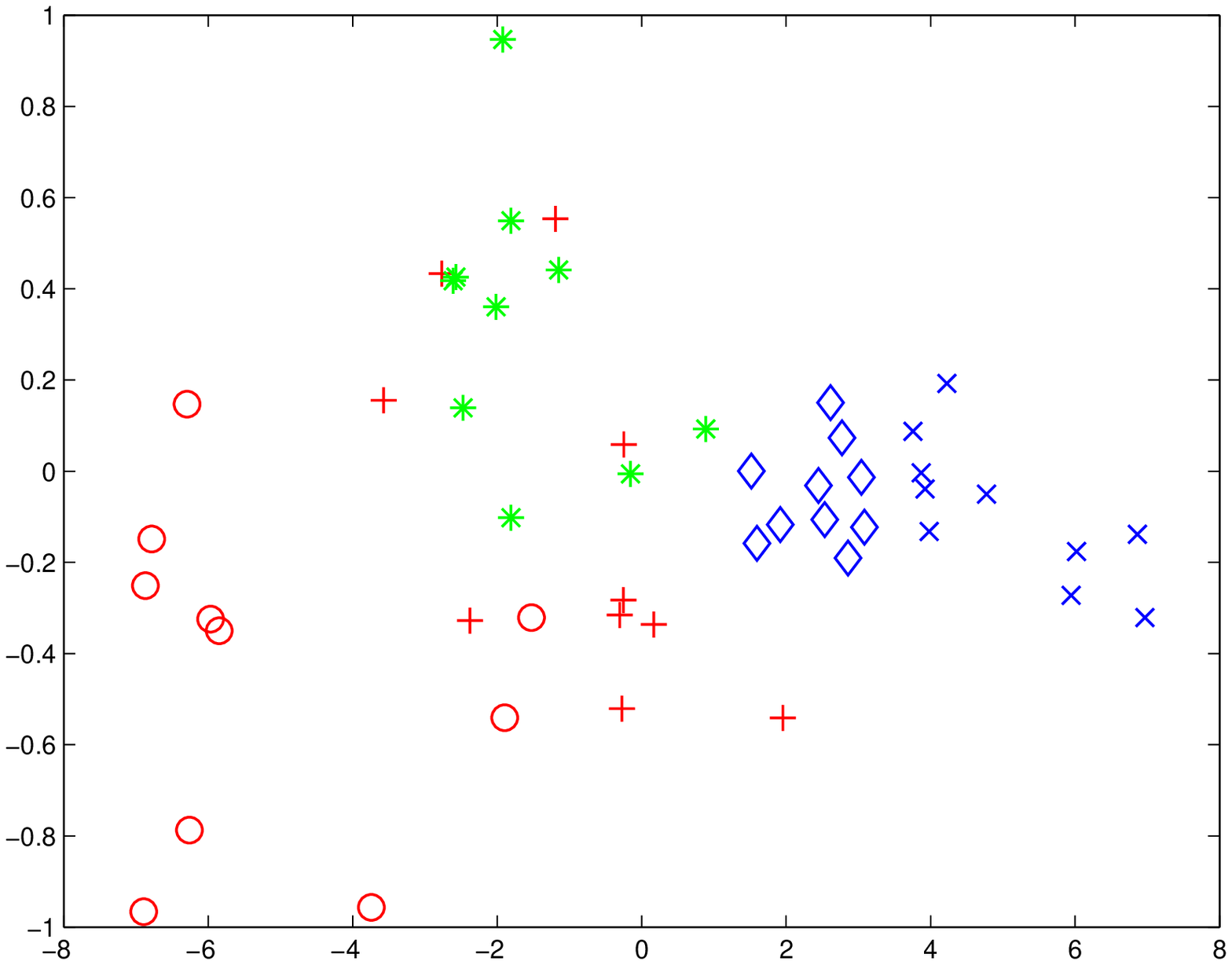}}\\
\subfigure[SSDR-MML Iteration 1]{\label{fig:toy_3}
\includegraphics*[width=0.32\textwidth]{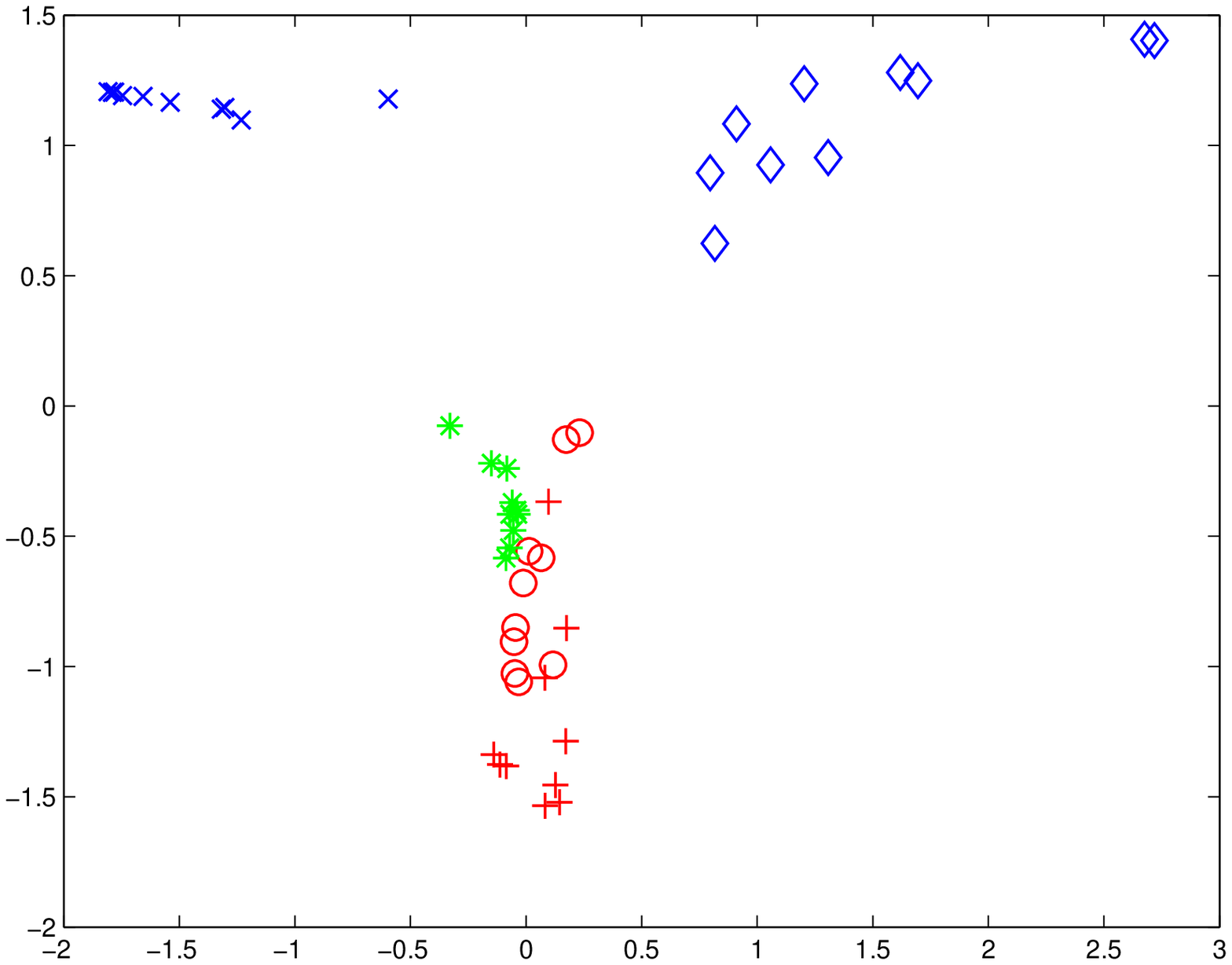}}
\subfigure[SSDR-MML Iteration 2]{\label{fig:toy_4}
\includegraphics*[width=0.32\textwidth]{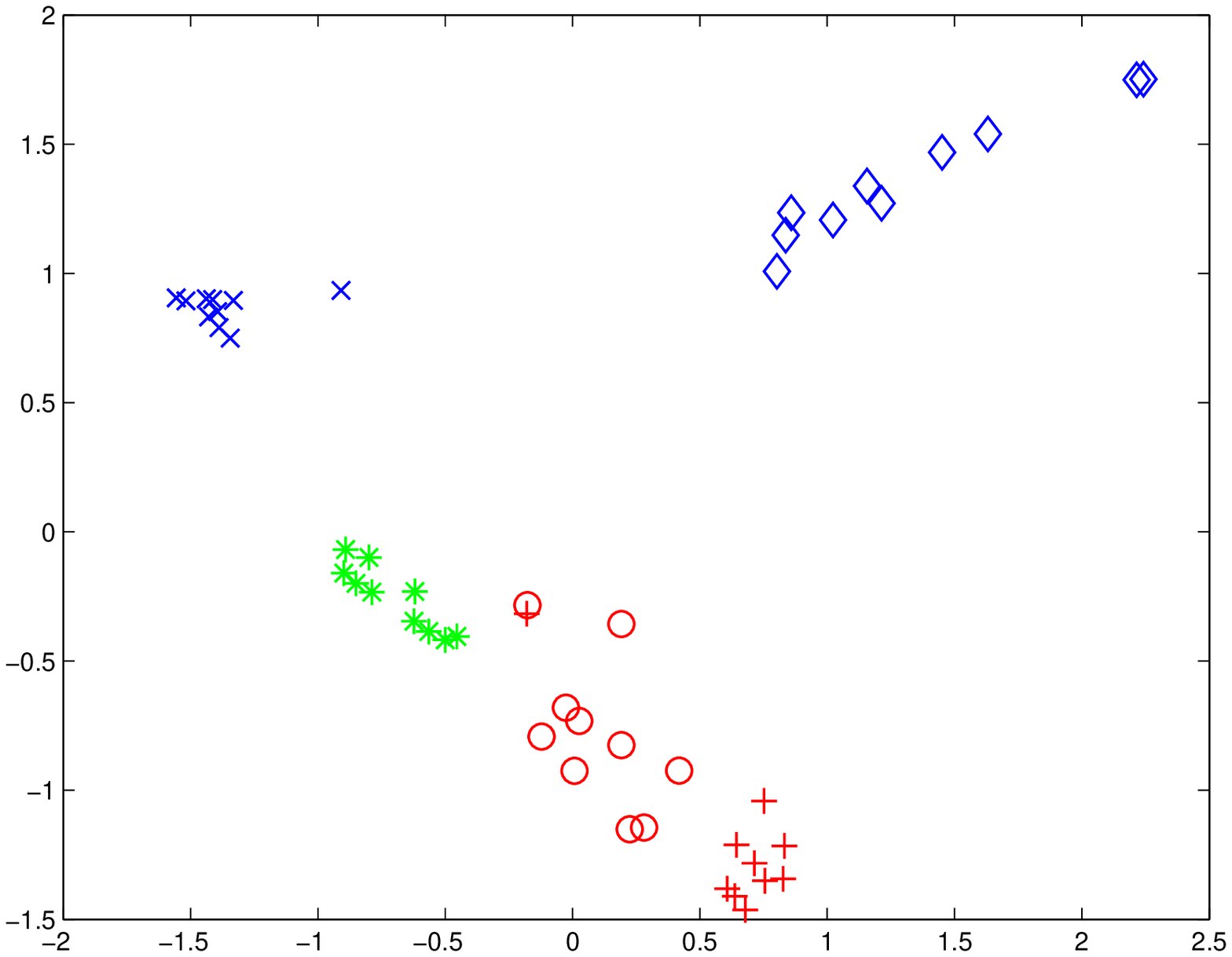}}
\subfigure[SSDR-MML Iteration 3]{\label{fig:toy_5}
\includegraphics*[width=0.32\textwidth]{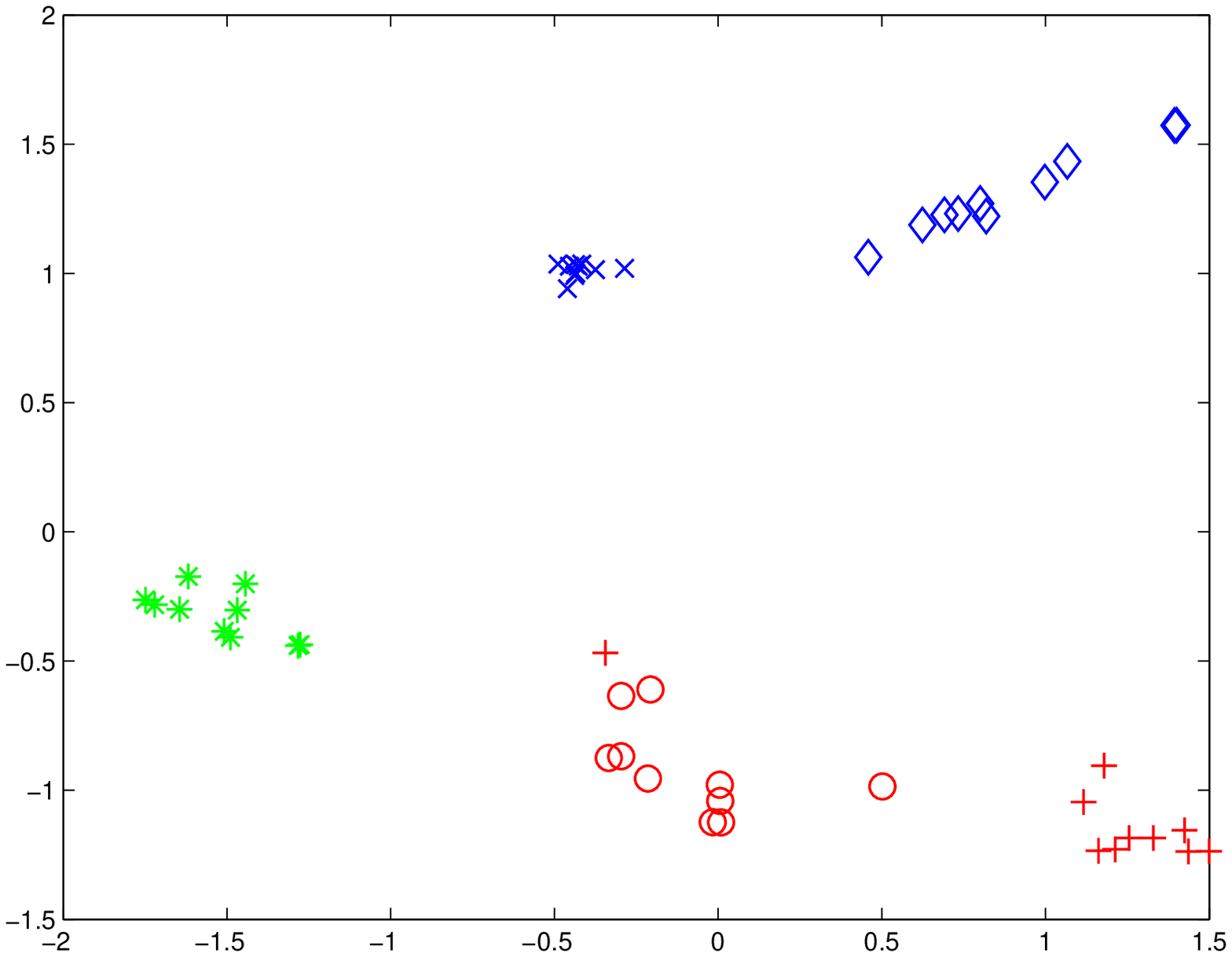}}
\caption{2D projected faces from different methods. Symbols denote different people, and colors denote different attributes.}\label{fig:toy}
\end{center}
\end{figure*}

Consider this simple experiment to illustrate the weakness of solving each problem independently. We collect $50$ frontal and well aligned face images of five people in ten different expressions, each of which are associated with four attributes: name, sex, bearded, glasses (see Fig.~\ref{fig:faces}). We shall project the face images into a 2D space by different techniques that perform unsupervised dimension reduction, supervised dimension reduction and finally our approach that simultaneously performs semi-supervised learning and dimension reduction for multi-label data. Fig.~\ref{fig:toy} shows these results where the five symbols denote five people and the three colors indicate the attributes associated to the face images. ``Red" stands for female, unbearded, and non-glasses; ``green" denotes male, unbearded, and non-glasses; and ``blue" indicates male, bearded, glasses. For Principal Component Analysis (PCA) \cite{pca}, an unsupervised dimension reduction technique, we see that it finds a mapping of the images into a 2D space (Fig.~\ref{fig:toy_1}) where the people are not well separated. This result is not surprising given PCA does not make use of the labeled data. Using the labels for only $30\%$ of images for supervision, we see that PCA+LDA \cite{fisher} performs only marginally better in Fig.~\ref{fig:toy_2} because the missing labels can not be inferred. Our approach simultaneously imputes/propagates the missing labels and performs dimension reduction for this multi-label data and, as shown in Fig.~\ref{fig:toy_3} to~\ref{fig:toy_5}, produces accurate predictions and monotonic improvement. During the iterative process, images belonging to the same person or associated with similar attributes aggregate gradually while dissimilar images move further apart. 

A subsequent risk in applying learning techniques to these challenging environments is that it is difficult to know if the learning approach was successful. Elegant frameworks such as structural and empirical risk minimization, though useful, have not been well extended and applied multi-label and multi-view settings. More empirical evaluation approaches such as cross-fold validation do not work well in sparse label settings. In this work we explore explicitly modeling the mechanism to combine multiple tasks and views as a data structure such that we can more clearly see how the labels are propagated and the relationship between tasks and views. By examining properties of this structure we can determine the success of the learning approach.

Our proposed work makes several contributions. We create a reconstruction error based framework that can model and create classifiers for complex data sets that: contain many missing labels, and are multi-view and multi-task/label. Multi-view learning involves multiple feature sets for the same set of objects. Previous work \cite{cotrain,Yarowsky95} shows that simply placing all the feature sets into one single view is suboptimal and raises difficult engineering issues if the views are fundamentally different (i.e. binary indicators in one view and real values in the other, or dimension and normalization issues). 
Our reconstruction error framework makes the following contributions to the field: 
\begin{itemize}
\item
Simultaneously perform dimension reduction, multi-task/label propagation in a multi-view and semi-supervised setting. 
\item
Explicitly model and construct a sparse graph showing which points are related.
\item
Allow the domain experts to clearly understand where/how the labels were propagated and how the views are complimentary using the above mentioned graph.
\item
Allow multi-view learning setting without the requirements of co-training: conditional independence of views and that each view (by itself) is sufficient for building a weak classifier. This is achieved since we do not serialize the classification learning problem instead learning from both views simultaneously.
\item
Allow a quantification of how successful the learning process was by examining the properties of the graph mentioned above (see Section \ref{sec:success}).
\end{itemize}

We begin our paper with Section \ref{sec:earlier} that overviews our earlier \cite{ssdrmc} which only considered the multi-label setting. We extend this to a more general framework in Section \ref{sec:general}. Sections \ref{sec:mtl} and \ref{sec:mvl} show specific formulations for multi-task and multi-view learning along with the optimization algorithm we use for each.  Section \ref{sec:spectral} shows how we can perform dimension reduction using our approach while Section \ref{sec:success} presents new work and that describes how to examine the data structure our framework uses to determine how successful the approach was. The experimental results in this section are particularly promising. Section \ref{sec:imple} discusses implementation issues and their solutions. Our experimental section (Section \ref{sec:experiment}) shows results that compare this work against existing competing techniques. The new experiments include comparing against a larger set of competing algorithms and experiments verifying the usefulness of our measure of how well the learnt model performs. For multi-task learning we show the strong result that as the number of labeled points increase our approach monotonically improves and outperforms four other approaches (see Fig.~\ref{fig:multitask_result}). In the multi-view setting our approach significantly outperforms all other approaches 13 out of 17 experiments on Caltech-256 image dataset and a set of UCI benchmarks. In Section \ref{sec:rework} we discuss related work and conclude our work in Section \ref{sec:conclu}.

\textbf{Differences to Conference Version.} The additional work that is in this paper and not the conference version is: (1) Extensions to our framework to handle both multi-task and multi-view learning settings, (2) Adding in a node regularizer to facilitate the learning with imbalanced labeling, (3) Proposing a measure to quantify the success of multi-task and multi-view learning, and (4) Extensive discussions and experiments.

We claim that our approach:
\begin{itemize}
\item
Has a well defined objective function (see Eq.~(\ref{eq:rekt1})) and is efficiently solvable using an iterative algorithm (see Section \ref{sec:alter_tasks} and \ref{sec:alter_views}).
\item
Produces a simpler model via the introduction of the sparsity requirement that prevents excessive label propagation (compare columns ``SSDR-MML nonsparse" and ``SSDR-MML" in Table \ref{tab:caltech_results} and \ref{tab:uci_results}).
\item
Performs better than simply placing multiple views into single view equivalent (compare columns ``SSDR-MML Simple" and ``SSDR-MML" in Table \ref{tab:caltech_results} and \ref{tab:uci_results}).
\item
Presents an alternating optimization that gradually propagates the labels and produces more reliable results (see variances in Fig.~\ref{fig:multitask_result}, Table \ref{tab:caltech_results} and \ref{tab:uci_results}).
\end{itemize}

\section{Earlier Framework} \label{sec:earlier}

Our earlier work \cite{ssdrmc} aimed to simultaneously solve the transductive inference and dimension reduction problems for just multi-label learning. For this task, a reasonable choice of the cost function is reconstruction error \cite{lle}, which attempts to discover nonlinear structure \textbf{amongst the features} of the high dimensional data by exploiting a \textbf{linear combination} of instances. In this way, each feature vector (row vector $\mathbf{x}_i$) and label vector (row vector $\mathbf{f}_i$) can be represented by a weighted linear combination of the corresponding $k$ nearest neighbor vectors ($\mathcal{N}_i$), and the problem is to optimize the weight matrix $\mathbf{W}$ and classifying matrix $\mathbf{F}$ (which contains the instance labels) simultaneously. As in earlier semi-supervised work \cite{gfhf}, we will abuse notation also refer to $\mathbf{F}$ as a function. Note since we explore a semi-supervised setting $\mathbf{F}$ may be partially filled in with given labels. Thus we can formulate the objective function as:

\begin{equation} \label{eq:ssdr-mc1}
\mathcal{Q}(\mathbf{W},\mathbf{F})=(1-\alpha)\sum_{i=1}^n\|\mathbf{x}_i-\sum_{j\in\mathcal{N}_i}\mathbf{W}_{ij}\mathbf{x}_j\|_\mathcal{F}^2+\alpha\sum_{i=1}^n\|\mathbf{f}_i-\sum_{j\in\mathcal{N}_i}\mathbf{W}_{ij}\mathbf{f}_j\|_\mathcal{F}^2
\end{equation}\\
where the first term is the feature reconstruction error which measures the error between the position of a point written as a linear combination of its nearest neighbors. The second term is the label reconstruction error, which measures the error of writing the labels of a point as a linear combination of the labels of its nearest neighbors. The tuning parameter $0\leq \alpha\leq 1$ determines how much the weights should be biased by the labels. It is advantageous to learn the weight matrix $\mathbf{W}$ as it is invariant to \emph{rotations} and \emph{rescalings} \cite{lle}, which follows immediately from the form of Eq.~(\ref{eq:ssdr-mc1}). In order to guarantee the invariance to \emph{translations}, we enforce the sum-to-one constraint upon each row of $\mathbf{W}$. We use the same weight matrix $\mathbf{W}$ for both feature vectors and label vectors, and constrain the prior/given labels to be unchangeable ($\mathbf{F}_l=\mathbf{Y}$). The optimal $\mathbf{W}$ and $\mathbf{F}$ matrix can be obtained by minimizing the reconstruction error function, and thus our problem can be expressed as a constrained optimization:

\begin{eqnarray} \label{eq:ssdr-mc2}
&&\min\,\,\mathcal{Q}(\mathbf{W},\mathbf{F}) \nonumber \\
&&s.t.\quad\,\,\,\,\mathbf{F}_l=\mathbf{Y} \nonumber \\ &&\quad
\sum_{j\in\mathcal{N}_i}\mathbf{W}_{ij}=1,\,i=1,\cdots,n.
\end{eqnarray}

\section{Extended Formulation}  \label{sec:general}

Here we extend the formulation in Eq.~(\ref{eq:ssdr-mc2}) to allow multiple views (feature vectors) and tasks (label vectors). We draw a distinction between multi-label and multi-task learning as follows. In the former each instance can have multiple binary labels that only take on a \textbf{binary} value (for instance in our example in Fig.~\ref{fig:toy} each image has three binary labels, i.e. bearded, glasses and gender) while in the latter, each instance can be involved in multiple multi-class tasks each of which can take on a \textbf{discrete} value within a finite label set.

\textbf{Notation.} Given a set of $n$ instances containing both labeled and unlabeled data points, we define a general learning problem on multiple tasks and multiple feature spaces. In the learning problem, there are $p$ related tasks $\mathcal{T} = \lbrace t_1, t_2, \cdots, t_p\rbrace$, each of which can be a multi-class task with a given finite label set. For the $k$-th task $t_k$, we define a binary classifying function $\mathbf{F}^k\in\mathbb{B}^{n\times c^k}$ on its corresponding label set $\mathcal{C}^k = \lbrace 1, 2, \cdots, c^k\rbrace$. Note that each instance will have $p$ label vectors $\lbrace \mathbf{f}^1 \ldots \mathbf{f}^k \rbrace$, containing the label sets for each task. Then $f_{ij}^k=1$ if and only if the $i^{th}$ instance for task $k$ has the label $j^{th}$ label for that task (the $i^{th}$ instance $\in$ class $\mathcal{C}^k_j$) and $f_{ij}^k=0$ otherwise. Without loss of generality, we assume that the points have been reordered so that the first $l^k$ points for task $k$ are labeled with the remaining $u^k$ points being unlabeled (where $n=l^k+u^k$ and typically $l^k\ll n$), and construct a prior label matrix $\mathbf{Y}^{k} \in \mathbb{B}^{l^k \times c^k}$ using the given labels in task $t_k$. Similarly, for each instance, there are $q$ feature descriptions from different views $\lbrace \mathbf{x}^1,\cdots,\mathbf{x}^q \rbrace$ with $\mathbf{x}_i^k$ being the $k^{th}$ feature description/view of the $i^{th}$ instance. Our aim is to create an asymmetric graph $G(V,E)$ over the $n$ instances represented by the weight matrix $\mathbf{W}\in\mathbb{R}^{n\times n}$, where we set $w_{ii}=0$ to avoid self-reinforcement. The diagonal node degree matrix is denoted by $\mathbf{D}$, where $d_{jj}=\sum_{i=1}^n w_{ij}$. This article will often refer to row and column vectors of matrices, for example, the $i$-th row and $j$-th column vectors of $\mathbf{W}$ are denoted as $\mathbf{w}_{i\bullet}$ and $\mathbf{w}_{\bullet j}$, respectively. The notations are summarized in Table~\ref{tab:notation}.

\begin{table}[!ht] 
\begin{center}
\caption{Notation Table} \label{tab:notation}
\begin{tabular}{|c||c|}
  \hline
Notation & Description \\
  \hline
  \hline
$t_i$ & The $i^{th}$ task \\
  \hline
$\mathbf{F}^k, \mathbf{f}_{i\bullet}^k$ & Binary classifying matrix (vector) for $k^{th}$ task ($i^{th}$ instance) \\  
  \hline
$\mathbf{Y}^k, \mathbf{y}_{i\bullet}^k$ & Given label matrix (vector) for $k^{th}$ task ($i^{th}$ instance) \\  
  \hline
$\mathbf{x}_i^k$ & Feature description of $i^{th}$ instance for $k^{th}$ view  \\  
  \hline
$\mathbf{W}, \mathbf{w}_{i\bullet}$ & Graph weight matrix (vector, $i^{th} row$) \\  
  \hline
$\mathbf{D}$ & Graph degree matrix \\ 
  \hline  
$\mathbf{V}^k$ & Node regularizer for $k^{th}$ task\\
  \hline
$C^{\mathbf{z}}$ & Local covariance matrix of a vector $\mathbf{z}$ \\
  \hline
$n, p, q$ & Number of instances, tasks, views \\
  \hline
$l^k, u^k$ & Number of labeled (unlabeled) instances in $k^{th}$ task \\
  \hline
$I$ & Identity matrix \\
  \hline
$\mathbf{1}, \mathbf{0}$ & Vector with all ones (zeros) entries \\
  \hline
$\alpha_k,\beta_k, \lambda$ &  Tuning parameters\\ 
  \hline 
\end{tabular}
\end{center}
\end{table}

\textbf{The Framework.} The key question in multi-task or multi-view learning is how to incorporate the information carried by related tasks or feature spaces into the same learning problem. In graph transduction, such kind of information encoding can be expressed in terms of partially fitting the graph to all available tasks or views based on their importance or relatedness. As before, we solve the label inference problem by following the intuition that ``nearby" points tend to have similar labels, and adopt \emph{reconstruction error} \cite{lle,ssdrmc}. These nearby points are learnt by our algorithm and encoded in $\mathbf{W}$, and we can then use $\mathbf{W}$ to propagate the labeled points to their neighbors. To ensure that no given labels are over-written we constrain $\mathbf{F}_l^k=\mathbf{Y}^k$, that is the given labels for a task are never overwritten. Finally, to ensure that the labels are not over-propagated (something we later test) we add in a regularization term to enforce the graph sparsity. The general learning framework can then be formulated as:

\begin{eqnarray} \label{eq:rekt1}
\mathcal{Q}(\mathbf{W},\mathbf{F})&=&\sum_{k=1}^q\alpha_k\sum_{i=1}^n \|\mathbf{x}^{k}_i-\sum_{j=1}^n w_{ij}\mathbf{x}^{k}_j\|_\mathcal{F}^2+\sum_{k=1}^{p}\beta_k\sum_{i=1}^n\|\mathbf{f}^{k}_{i\bullet}-\sum_{j=1}^n w_{ij}\mathbf{f}^{k}_{j\bullet}\|_\mathcal{F}^2+ \lambda\| \mathbf{W}\|_\mathcal{F}^2 \nonumber \\
\lefteqn{\mathrm{s.t.}\quad \forall i,~ \mathbf{w}_{i\bullet}\mathbf{1}=1;\quad 
\mathbf{F}^k_l=\mathbf{V}^k\mathbf{Y}^k.}
\end{eqnarray}\\
where $\|\bullet\|_\mathcal{F}$ denotes the Frobenius norm. The first term in Eq.~\ref{eq:rekt1} is the reconstruction error over the multiple feature descriptions of instances, the second term is the reconstruction error over the classifying functions for the multiple tasks, and the third term is the $\mathcal{L}_2$ penalty to enforce graph sparsity. Note that the two reconstruction error terms share exactly the \emph{same} weight matrix $\mathbf{W}$, which enables the multiple tasks and views to help each other. The tuning parameter $0\leq\alpha_k\leq 1$ is determined by the ``importance" of feature descriptions at the $k$-th view, and $0\leq\beta_k\leq 1$ is decided by the ``relatedness" between the task $t_k$ to other tasks. $\lambda$ controls the sparsity. We will in Section~\ref{sec:success} provide a success measure to guide the selection of these parameters. The objective function consists of the reconstruction errors of feature spaces at different views and multiple related tasks, as allowing the graph weight partially fits to each of them. 

\textbf{Overcoming Class Imbalance.} If one class is more popular than the another, there is the chance that even though the labels of the less frequent class are propagated they are ignored in favor of the popular class. To overcome this, 
we introduce the matrix $\mathbf{V}^k$ which is a node regularizer \cite{gtam} to balance the influence of different class labels in task $t_k$. The matrix $\mathbf{V}^k = \mathrm{diag}(\mathbf{v}^k)$ is a function of the labels for the $k^{th}$ task, $\mathbf{Y}^k$ and $\mathbf{D}_l$ is the degree matrix of the labeled points calculated from $\mathbf{W}$:

\begin{equation} \label{eq:rekt2}
\mathbf{v}^k = \sum_{i=1}^{c^k}\dfrac{\mathbf{y}^k_{\bullet i}\odot \mathbf{D}_l\mathbf{1}}{\left({\mathbf{y}^k_{\bullet i}}\right)^T \mathbf{D}_l\mathbf{1}}
\end{equation} \\
where the symbol $\odot$ denotes Hadamard product (element-wise multiplication) and $\mathbf{1}=[ 1, 1,\cdots, 1 ]^T$. By definition, $\mathbf{V}^k\mathbf{Y}^k$ is a normalized version of the label matrix $\mathbf{Y}^k$ satisfying $\sum_i (\mathbf{V}^k\mathbf{Y}^k)_{ij} = 1$ for $\forall j$. The normalized label matrix $\mathbf{V}^k\mathbf{Y}^k$ empowers the highly connected instances to contribute more during the graph diffusion and label propagation process. Since the total diffusion of each class is normalized to one, the influence of different classes is balanced even if the given labels are imbalanced, and equally balanced class generally leads to more reliable solutions. In Section \ref{sec:mtl} and \ref{sec:mvl}, we will explicitly apply the general framework to multi-task and multi-view learning problems, respectively.

\textbf{Benefits.} The proposed SSDR-MML is a general learning framework which has several key benefits. We list and describe these benefits below and will examine them later in this paper.
\begin{itemize}
\item
In Section \ref{sec:experiment}, we shall examine and experimentally verify multi-task and multi-view setting separately, but we can easily apply our work to multi-task and multi-view learning. This allows us to discover both the correlation among multiple tasks and the complementary aspects between multiple feature descriptions to improve the performance of the learning task at hand. As the weight matrix $\mathbf{W}$ is partially fitted to each task and view, the aforementioned correlation and complementarity are incorporated into the learning mechanism, and thus enables the learner to take advantage of the multiple knowledge sources. 
\item
SSDR-MML is not limited to the two basic situations (multi-task and multi-view); it may be extended to handle other useful learning settings. For example, it can be reformulated as an online multi-task or multi-view learning algorithm since the learning of the weights for each instance $\mathbf{w}_{i\bullet}$ is independent of others. 
\item
The output $\mathbf{W}$ completely specifies a set of lower dimensional representations for the original data, which can be obtained by spectral embedding of the reconstruction error \cite{lle}. We discuss this issue in Section \ref{sec:spectral}.
\item
Unlike many co-training algorithms, we do not make the conditional independence assumption among the multiple views. While knowledge combining in many multiple kernel machines is conducted in linear fashion, our SSDR-MML approach performs such kind of fusion using nonlinear method such that enables more detailed knowledge representations. 
\item
An important contribution of this work is a measure to quantify the success of learning, which can be used to determine whether a multi-task or multi-view learning was constructive. It also could be used to avoid the possible negative impacts of learning with multiple tasks or views, and may stimulate the development of active multi-task and multi-view learning algorithms.
\item
Our work can be naturally interpreted with the inference step being viewed as a process during which the labeled data points propagate their labels along the weighted edges to their neighbors. The learning of the matrix $\mathbf{W}$ similarly has the natural interpretation of learning the nearest neighbor graph that is most probable given the current labels. 
\end{itemize}

\textbf{Limitations.} Our proposed SSDR-MML framework can suffer from two major limitations: (1) potential sensitivity to parameter selection and how to chose the parameters; and (2) the objective function in Eq.~(\ref{eq:rekt1}) is not convex and hence the iterative algorithm we derived will only converge to a local optima. There is greater chance that destructive combining of multiple tasks or views may occur in our formulation since the objective function involves $p+q+1$ parameters in total. However, in practice we found that our framework is not sensitive to the regularization parameters (as shown in Fig.~\ref{fig:para}), and thereby the selection problem of $\alpha_k$ and $\beta_k$ is easily solvable under the guidance of the proposed success measure, as demonstrated in Fig.~\ref{fig:fwf}. While our objective function is non-convex in general, empirical results comparing our approach to other algorithms (some of which do converge to their global optimum) indicate that the additional complexity of our model is warranted and is pragmatically not a major concern. In future work, we shall explore making uses of the multiple local minima each of which corresponds to a particular interpretation of the data. 

We shall now describe specific solutions for our general formulation for the multi-task and multi-view settings. The generalized algorithm for both these settings and for multi-task-multi-view learning is shown in Table \ref{tab:tab1}.

\section{Multi-task Learning} \label{sec:mtl}

\subsection{Formulation}

We start by studying multi-task learning, where there are multiple learning tasks over the same set of feature descriptions of instances. Our motivation is to improve the learning performance on the multiple tasks by making use of the commonality among them. Intuitively, multi-task learning could greatly improve the learning performance if the multiple tasks are highly correlated. On the contrary, if there is no relatedness between the multiple tasks, multi-task learning cannot be beneficial and even could be detrimental. We can better understand this premise by interpreting our work as label propagation. For multi-task learning to be successful, points \emph{labeled with the first task} should have this label transfered/propagated to points \emph{labeled with other tasks} and vice-versa. If this propagation is extensive then the approach will be successful. This is the idea behind the math in our work on successfully identifying the success of our approach in section \ref{sec:success}. Under multi-task setting, the general framework shown in Eq.~(\ref{eq:rekt1}) can be reduced to:

\begin{eqnarray} \label{eq:rekt3}
\mathcal{Q}(\mathbf{W},\mathbf{F})&=& \alpha\sum_{i=1}^n\|\mathbf{x}_i-\sum_{j=1}^n w_{ij}\mathbf{x}_j\|_\mathcal{F}^2 + \sum_{k=1}^{p}\beta_k\sum_{i=1}^n\|\mathbf{f}^{k}_{i\bullet}-\sum_{j=1}^n w_{ij}\mathbf{f}^{k}_{j\bullet}\|_\mathcal{F}^2 + \lambda\| \mathbf{W}\|_\mathcal{F}^2 \nonumber \\
\lefteqn{\mathrm{s.t.}\quad \forall i,~ \mathbf{w}_{i\bullet}\mathbf{1}=1;\quad 
\mathbf{F}^k_l=\mathbf{V}^k\mathbf{Y}^k.}
\end{eqnarray}

\subsection{Alternating Optimization} \label{sec:alter_tasks}
The formulation shown in Eq.~(\ref{eq:rekt3}) is a minimization problem involving two variables to optimize. Since this objective is not convex it is difficult to simultaneously recover both unknowns. However, if we hold one unknown constant and solve the objective for the other, we have two convex problems that can be optimally solved in closed form. In the rest of this section, we propose an alternating optimization for the SSDR-MML framework, which iterates between the updates of $\mathbf{W}$ and $\mathbf{F}^k$ until $\mathbf{F}^k$ stabilized. The experimental results (see Table~\ref{tab:caltech_results} and \ref{tab:uci_results}) indicate that converging to the local optima still provides good results and is better than less complex objective functions that are solved exactly.

\subsubsection{Update for $\mathbf{W}$} 
If the classifying function $\mathbf{F}^k$ is a constant, then the weight matrix $\mathbf{W}$ can be recovered in closed form as a constrained least square problem. Since the optimal weights for reconstructing a particular point is only dependent on other points, each row of the weight matrix $\mathbf{W}$ can be obtained independently. The problem reduces to minimize the following function:

\begin{eqnarray} \label{eq:ssdr-mc3}
\min_{\mathbf{w}_{i\bullet}}\quad \mathcal{Q}(\mathbf{w}_{i\bullet}) &=& \alpha\|\mathbf{x}_i - \mathbf{w}_{i\bullet} \mathbf{X}'_i\|_\mathcal{F}^2+\sum^{p}_{k=1}\beta^k \|\mathbf{f}^k_{i\bullet}-\mathbf{w}_{i\bullet} {\mathbf{F}^{k}_{i}}'\|_\mathcal{F}^2 +\lambda\|\mathbf{w}_{i\bullet} \|_\mathcal{F}^2 \nonumber \\
s.t.\quad\quad \mathbf{w}_{i\bullet}\mathbf{1}&=&1 
\end{eqnarray}\\
where $\mathbf{X}'_{i}$ and ${\mathbf{F}^{k}_{i}}'$ denote the set difference $\{\mathbf{X}\setminus \mathbf{x}_i\}$ and $\{\mathbf{F}^k\setminus \mathbf{f}^k_{i\bullet}\}$ respectively, i.e. the set of all instances and their labels except the $i^{th}$ instance and its labels, and as before $\|\bullet\|_\mathcal{F}$ denotes Frobenius norm. The derivative of the cost function with respect to $\mathbf{w}_{i\bullet}$ can be written as:

\begin{equation} \label{eq:ssdr-mc4}
\nabla_{\mathbf{w}_{i\bullet}}\mathcal{Q}(\mathbf{w}_{i\bullet}) = \mathbf{w}_{i\bullet} \left(\alpha \left(\mathbf{1}\mathbf{x}_i-\mathbf{X}'_{i}\right)\left(\mathbf{1}\mathbf{x}_i-\mathbf{X}'_{i}\right)^T + \sum_{k=1}^p\beta^k \left(\mathbf{1}\mathbf{f}^k_{i\bullet}-{\mathbf{F}^{k}_{i}}'\right)\left(\mathbf{1}\mathbf{f}^k_{i\bullet}-{\mathbf{F}^{k}_{i}}'\right)^T + \lambda I\right) 
\end{equation}

To provide the solution for $\mathbf{W}$, we first introduce the local covariance matrix. Let $C^{\mathbf{x}_i}$ denote the local covariance matrix of the feature description $\mathbf{x}_i$ of the $i^{th}$ instance. The term ``local" refers to the fact that the instance is used as the mean of the calculation.

\begin{equation} \label{eq:rekt4}
C^{\mathbf{x}_i} = \left(\mathbf{1}\mathbf{x}_i-\mathbf{X}'_{i}\right)\left(\mathbf{1}\mathbf{x}_i-\mathbf{X}'_{i}\right)^T
\end{equation} \\
where as before $\mathbf{x}_i$ is a row vector, and $\mathbf{1}$ is a column vector with all one entries. Using a Lagrange multiplier to enforce the sum-to-one constraint, the update of $\mathbf{w}_{i\bullet}$ (the weights for the $i^{th}$ instance) can be expressed in terms of the inverse local covariance matrices.

\begin{equation} \label{eq:rekt5}
\mathbf{w}_{i\bullet}=\dfrac{\mathbf{1}^T\left(\alpha C^{\mathbf{x}_i} + \sum_{k=1}^p\beta^k C^{\mathbf{f}^k_{i\bullet}} + \lambda I\right)^{-1}}
{\mathbf{1}^T\left(\alpha C^{\mathbf{x}_i} + \sum_{k=1}^p\beta^k C^{\mathbf{f}^k_{i\bullet}} + \lambda I\right)^{-1}\mathbf{1}}
\end{equation} \\
where $C^{\mathbf{f}^k_{i\bullet}}$ is defined in the same manner as $C^{\mathbf{x}_i}$ shown in Eq.~(\ref{eq:rekt4}), and $I$ represents the identity matrix. As previously defined, $\alpha$ and $\beta$ are the regularization parameters for features and labels respectively, and $\lambda$ controls the sparsity. As we obtained $\mathbf{w}_{i\bullet}$ of all $n$ instances $i=1,\cdots,n$, the optimal weight matrix $\mathbf{W}$ can be constructed by simply placing each weight vector to its corresponding location in the matrix. Notice that Eq.~(\ref{eq:rekt5}) can not guarantee the weights are non-negative. We empirically found that the negative weights are infrequent and relatively small, and thus only have little effect to the learning performance. In the proposed framework, we keep the negative weights since positive weights indicate two points are similar then negative weights indicate two points are dissimilar. Intuitively, if a positive weight means that the corresponding neighbor constructively contributes to the reconstruction, a negative weight indicates a negative contribution. 

\subsubsection{Update for $\mathbf{F}^k$} \label{sec:updatef}
Here we assume the weight matrix $\mathbf{W}$ is constant, then the goal is to fill in the missing labels $\mathbf{F}_u^k$ based on the fixed weight matrix $\mathbf{W}$. For each learning task $t_k$, we relax the binary classifying function $\mathbf{F}^k$ to be real-valued, so that the optimal $\mathbf{F}_u^k$ can be recovered in closed form. Since the feature reconstruction error (first term in Eq.~(\ref{eq:rekt3}) ) is a constant, we can rewrite the formulation in matrix format:

\begin{eqnarray} \label{eq:rekt6}
&&\min_{\mathbf{F}^k}\quad \mathcal{Q}(\mathbf{F}^k)=\dfrac{1}{2}tr\left\lbrace \left({\mathbf{F}^k}\right)^T\left(I-\mathbf{W}\right)^T\left(I-\mathbf{W}\right)\mathbf{F}^k\right\rbrace \nonumber \\
&&s.t.\quad\quad \mathbf{F}^k_l=\mathbf{V}^k\mathbf{Y}^k
\end{eqnarray}\\
where $\mathbf{Y}^k$ carries the given labels in the $k^{th}$ task, and $\mathbf{V}^k$ is the corresponding the node regularizer. To express the solution in terms of matrix operations, we assume the instances have been ordered so that the first $l$ are labeled and the remaining $u$ are the unlabeled instances. We can then split the weight matrix $\mathbf{W}$ and classification function $\mathbf{F}^k$ after the $l^k$th row and column, i.e. $\mathbf{W}=\left[\begin{array}{ll}\mathbf{W}_{ll}\,\,\,\mathbf{W}_{lu}\\ \mathbf{W}_{ul}\,\,\mathbf{W}_{uu} \end{array}\right]$ and $\mathbf{F}^k=\left[\begin{array}{ll}\mathbf{F}^k_{l}\\\mathbf{F}^k_{u}\end{array}\right]$. Note that we do not attempt to overwrite the labeled instances. The cost function is convex, thereby allowing us to recover the optimal $\mathbf{F}^k$ by setting the derivative $\nabla_{\mathbf{F}^k}\mathcal{Q}(\mathbf{F}^k) = 0$.

\begin{eqnarray} \label{eq:ssdr-mc6}
&&\left(I-\left[\begin{array}{ll}\mathbf{W}_{ll}\,\,\,\mathbf{W}_{lu}\\ \mathbf{W}_{ul}\,\,\mathbf{W}_{uu} \end{array}\right]\right)^T\left(I-\left[\begin{array}{ll}\mathbf{W}_{ll}\,\,\,\mathbf{W}_{lu}\\ \mathbf{W}_{ul}\,\,\mathbf{W}_{uu} \end{array}\right]\right)\left[\begin{array}{ll}\mathbf{F}^k_{l}\\\mathbf{F}^k_{u}\end{array}\right]^k=\mathbf{0} \\ \nonumber
\\\nonumber
&&\mathrm{s.t.}\quad\quad\quad \mathbf{F}^k_l= \mathbf{V}^k\mathbf{Y}^k.
\end{eqnarray}\\
where $\mathbf{0}$ is a matrix with all zero entries. The optimization problem above yields a large, sparse system of linear equations that can be solved by a number of standard methods. The most straightforward one is the closed-form solution via matrix inversion. The prediction for the unlabeled instances can be obtained in closed form via matrix inversion:

\begin{eqnarray} \label{eq:rekt7}
\mathbf{F}^k_u =\left(\mathbf{W}_{lu}^T\mathbf{W}_{lu}+\left(I_u-\mathbf{W}_{uu}\right)^T\left(I_u-\mathbf{W}_{uu}\right)\right)^{-1}\left( \mathbf{W}_{lu}^T\left(I_l-\mathbf{W}_{ll}\right)+\left(I_u-\mathbf{W}_{uu}\right)^T\mathbf{W}_{ul}\right) \mathbf{V}^k\mathbf{Y}^k
\end{eqnarray}\\
where $I_l$ and $I_u$ denote identity matrices with dimension $l$ and $u$, respectively. The final predictions for an unlabeled instances $\mathbf{I}_i$ can be obtained by set $f^k_{ij}=1$ where $j=\mathrm{arg\,max}_j~f^k_{ij}$, and set other elements in $\mathbf{f}^k_{i\bullet}$ to zeros. 

\subsubsection{Progressive Update for $\mathbf{Y}^k$} \label{sec:updatepf}
Since there is no theoretical convergence guarantee for the proposed alternating optimization, it is possible that the prediction of the current iteration oscillates and backtracks from the predicted labellings in previous iterations. A straightforward solution to address this problem is to set up a small tolerance, but it is difficult for a practitioner to set the value of tolerance. Alternatively, we propose a progressive method to update $\mathbf{Y}^k$ incrementally instead of updating $\mathbf{F}^k$, as to remove the backtrackings, inconsistency and unstable oscillations. In each iteration, we only make the most confident prediction, and treat this prediction as the ground truth in the future training. For this purpose, we also consider the prior label matrix $\mathbf{Y}^k$ as an unknown in the cost function Eq.~(\ref{eq:rekt6}). Choosing the most confident prediction is guided by the direction with largest negative gradient in the partial derivative $\dfrac{\partial{\mathcal{Q}(\mathbf{F}^k,\mathbf{Y}^k)}}{\partial{\mathbf{F}^k}}$.

\begin{equation} \label{eq:rekt8}
\dfrac{\partial{\mathcal{Q}}}{\partial{\mathbf{F}^k}} = \left[\begin{array}{ll}\left(\dfrac{\partial{\mathcal{Q}}}{\partial{\mathbf{F}^k}}\right)_l\\\left(\dfrac{\partial{\mathcal{Q}}}{\partial{\mathbf{F}^k}}\right)_u\end{array}\right] = (I-\mathbf{W})^T(I-\mathbf{W})\left[\begin{array}{ll}\mathbf{V}^k\mathbf{Y}^k\\\mathbf{0}\end{array}\right]
\end{equation}

The most confident prediction is located at position $(i^*,j^*)^k$, which can be decided by finding the largest negative value.

\begin{eqnarray} \label{eq:rekt9}
(i^*,j^*)^k = \mathrm{arg\,min}_{i,j} \left(\dfrac{\partial{\mathcal{Q}}}{\partial{\mathbf{F}^k}}\right)_u 
\end{eqnarray}

In $m$-th iteration, we locate the position $(i^*,j^*)^k$ in the matrix $\mathbf{F}^k_u$, and reset the values of entries in $\mathbf{f}^k_{(l^k+i^*)\bullet}$. Specifically, we set the entry $f^k_{l^k+i^*,j^*}$ to 1 and other entries in $(l_k+i^*)$-th row to 0, and then update $\mathbf{Y}^k$ by:

\begin{equation} \label{eq:rekt10}
\left({\mathbf{Y}^k}\right)^{m+1} = \left[\begin{array}{ll}\left({\mathbf{Y}^k}\right)^{m}\\\mathbf{f}^k_{\left(l^k+i^*\right)\bullet}\end{array}\right]
\end{equation}

After each prediction, we update $l^k$ with $l^k=l^k+1$, recompute the weight matrix ${\left(\mathbf{W}\right)}^{m+1}$ by Eq.~(\ref{eq:rekt5}) based on the newly obtained ${\left(\mathbf{Y}^k\right)}^{m+1}$, and update the node regularizer ${\left(\mathbf{V}\right)}^{m+1}$. The whole procedure repeats until the missing labels in all $p$ tasks have been predicted.

\section{Multi-view Learning} \label{sec:mvl}

\subsection{Formulation}
We now apply our framework to multi-view learning setting, where we assume that for the same learning task and the same set of instances, there are multiple feature descriptions obtained from different views. Here our goal is to improve learning performance by taking advantage of the complementary information carried by the multiple feature descriptions. Under this setting, the general framework shown in Eq.~(\ref{eq:rekt1}) can be simplified to:

\begin{eqnarray} \label{eq:rekt11}
\mathcal{Q}(\mathbf{W},\mathbf{F})&=&\sum_{k=1}^q\alpha_k\sum_{i=1}^n\|\mathbf{x}^{k}_i-\sum_{j=1}^n w_{ij}\mathbf{x}^{k}_j\|_\mathcal{F}^2+\beta\sum_{i=1}^n\|\mathbf{f}_{i\bullet}-\sum_{j=1}^n w_{ij}\mathbf{f}_{j\bullet}\|_\mathcal{F}^2 + \lambda\| \mathbf{W}\|_\mathcal{F}^2 \nonumber \\
\lefteqn{\mathrm{s.t.}\quad \forall i,~ \mathbf{w}_{i\bullet}\mathbf{1}=1;\quad 
\mathbf{F}_l=\mathbf{V}\mathbf{Y}.}
\end{eqnarray}

\subsection{Alternating Optimization} \label{sec:alter_views}
The optimization problem can be solved using the similar approach we proposed for multi-task setting, where a reliable local optima could be achieved by applying the proposed alternating optimization procedure shown in Section \ref{sec:alter_tasks}. 

\textbf{Update for $\mathbf{W}$.} While assuming the classification function $\mathbf{F}$ is fixed, the weight matrix $\mathbf{W}$ can be recovered as now a set of constrained least square problems. Given the definition of local covariance matrix shown in Eq.~(\ref{eq:rekt4}), the weights $\mathbf{w}_{i\bullet}$ for the $i^{th}$ instance can be solved independently by applying a Lagrange multiplier.

\begin{equation} \label{eq:rekt12}
\mathbf{w}_{i\bullet}=\frac{\mathbf{1}^T\left(\sum_{k=1}^q\alpha_k C^{\mathbf{x}^k_i} + \beta C^{\mathbf{f}_{i\bullet}} + \lambda I\right)^{-1}}
{\mathbf{1}^T\left(\sum_{k=1}^q\alpha_k C^{\mathbf{x}^k_i} + \beta C^{\mathbf{f}_{i\bullet}} + \lambda I\right)^{-1}\mathbf{1}}
\end{equation}

\textbf{Update for $\mathbf{F}$.} The classifying function $\mathbf{F}$ can be recovered by using the same methods proposed in Section~\ref{sec:updatef} or \ref{sec:updatepf}. Specifically, the update for $\mathbf{F}$ can be calculated using Eq.~(\ref{eq:rekt7}), or we can obtain the predictions using the progressive method shown from Eq.~(\ref{eq:rekt8}) to Eq.~(\ref{eq:rekt10}).

\section{Spectral Embedding for Dimension Reduction Step} \label{sec:spectral}
Here we describe an extension to the previous mentioned framework which allows dimension reduction to easily be performed. It can be used as a post-processing step after our algorithm has converged or as an additional step in our algorithm. It is useful as a method to more easily visualize the results of our algorithm (as done in Fig.~\ref{fig:toy_1}) or when working with high dimensional data (as done for the results in Fig.~\ref{fig:multitask_tmc}).

Since the weight matrix $\mathbf{W}$ completely captured the intrinsic geometric relations between data points, we can use it to perform dimension reduction. Note that the spectral embedding step is unnecessary for the learning purpose, but is only used to dimension reduction. Let $d$ denote the desired dimension of the feature vectors, the dimension reduced instance $\mathbf{\hat{x}}_i$ minimizes the embedding cost function:
\begin{equation} \label{eq:ssdr-mc8}
\mathcal{Q}(\mathbf{\hat{X}}) = \sum_{i=1}^n\|\mathbf{\hat{x}}_i-\sum_{j=1}^n\mathbf{W}_{ij}\mathbf{\hat{x}}_j\|^2
\end{equation}
where $\mathbf{\hat{X}}\in\mathbb{R}^{n\times d}$ is the dimension reduced data matrix. The embedding cost in Eq.~(\ref{eq:ssdr-mc8}) defines a quadratic form in the vector $\mathbf{\hat{x}}_i$. Since we want the problem well-posed and also to avoid trivial solutions, the minimization can be solved as a sparse eigen decomposition problem:
\begin{equation} \label{eq:ssdr-mc9}
\min\,\,\mathcal{Q}(\mathbf{\hat{X}}) = tr\left( \mathbf{\hat{X}}^T\mathbf{M}\mathbf{\hat{X}}\right)
\end{equation}
where $\mathbf{M}=(I-\mathbf{W})^T(I-\mathbf{W})$. The optimal embedding can be recovered by computing the smallest $d+1$ eigenvectors of the matrix $\mathbf{M}$, then discard the smallest eigenvector which is an unit vector. The remaining $d$ eigenvectors are the optimal embedding coordinates that minimize equation \ref{eq:ssdr-mc8}.

\section{Quantifying The Success of Multi-View and Multi-Task Learning} \label{sec:success}

Multi-task and Multi-view learning have been successfully applied to many real-world applications, however, in many data sets the performance is no better than and often worse than if each classification problem were solved independently \cite{transferornot}. Consequently, avoiding destructive fusing of information is an essential element in multi-task and multi-view learning. 
However, very few approaches produce a measure to determine if the transfer of knowledge between the views or tasks was successful. In this section we outline such a measure and in later empirically verify its usefulness.

\textbf{Multi-task Learning.} Our measure makes use of our interpretation of the SSDR-MML framework as performing label propagation and the mechanism for information combining $\mathbf{W}$. Let $\mathbf{F}$ denote a binary label matrix $\mathbf{F}\in \mathbb{B}^{n \times p}$, where $p$ is the number of tasks defined on a set of $n$ instances. $f_{ij} = 1$ if instance $i$ can be categorized into class $j$, and $f_{ij} = -1$ otherwise. After the training step we have available a weight matrix $\mathbf{W}$ built upon the set of instances, where $\mathbf{W}$ carries the information learnt from the multiple tasks and views. Since $\mathbf{W}$ row-wise sums to one, then we can use it as a random walk transition matrix. To quantify the success of multi-task learning, we define a measure of cross propagation (CP):

\begin{equation} \label{eq:rekt13}
\mathrm{CP}\left(\mathbf{W} \right) = \mathbf{F}^T {\left(\mathbf{W}\right)}^z \mathbf{F}
\end{equation}\\
where $z$ is a positive integer, indicating how many steps the labels have to propagate. The resulting $\mathrm{CP} \left(\mathbf{W} \right)$ is a $p \times p$ matrix where the entry at $i,j$ can be interpreted as \emph{how well the labels for task $i$ are propagated to the instances with labels for task $j$}. Therefore, the values on the diagonal measure the success of \emph{intra-task/label reconstruction}, and the off-diagonal values imply the success of \emph{inter-task/label reconstruction}. It has been widely reported \cite{multitask1,multitask2} that when task labels are correlated, multi-task learning performs best, then we can use the sum of all \textbf{off-diagonal} entries to indicate how well the knowledge transfer among multiple tasks has occurred. 

\textbf{Multi-vew Learning.} In a multi-view learning problem, there is only one task, thus $\mathrm{CP} \left(\mathbf{W} \right)$ is a single value measuring \emph{how well the class labels are reconstructed using the knowledge carried by multiple views}. For multi-view learning, a relatively large value in $\mathrm{CP} \left(\mathbf{W} \right)$ (compared to the ``single" case) implies that the joint learning of views was successful, while a smaller value indicates that detrimental view combining has occurred. The proposed success measure not only offers a way to quantify the performance of joint task or view learning, but also can be used to guide the selection of parameters in many existing multi-task or multi-view learning algorithms.

\begin{figure}[!ht]
\begin{center}
\subfigure[Multi-task learning on Yeast dataset]{\label{fig:fwf_1}
\includegraphics*[width=0.46\textwidth]{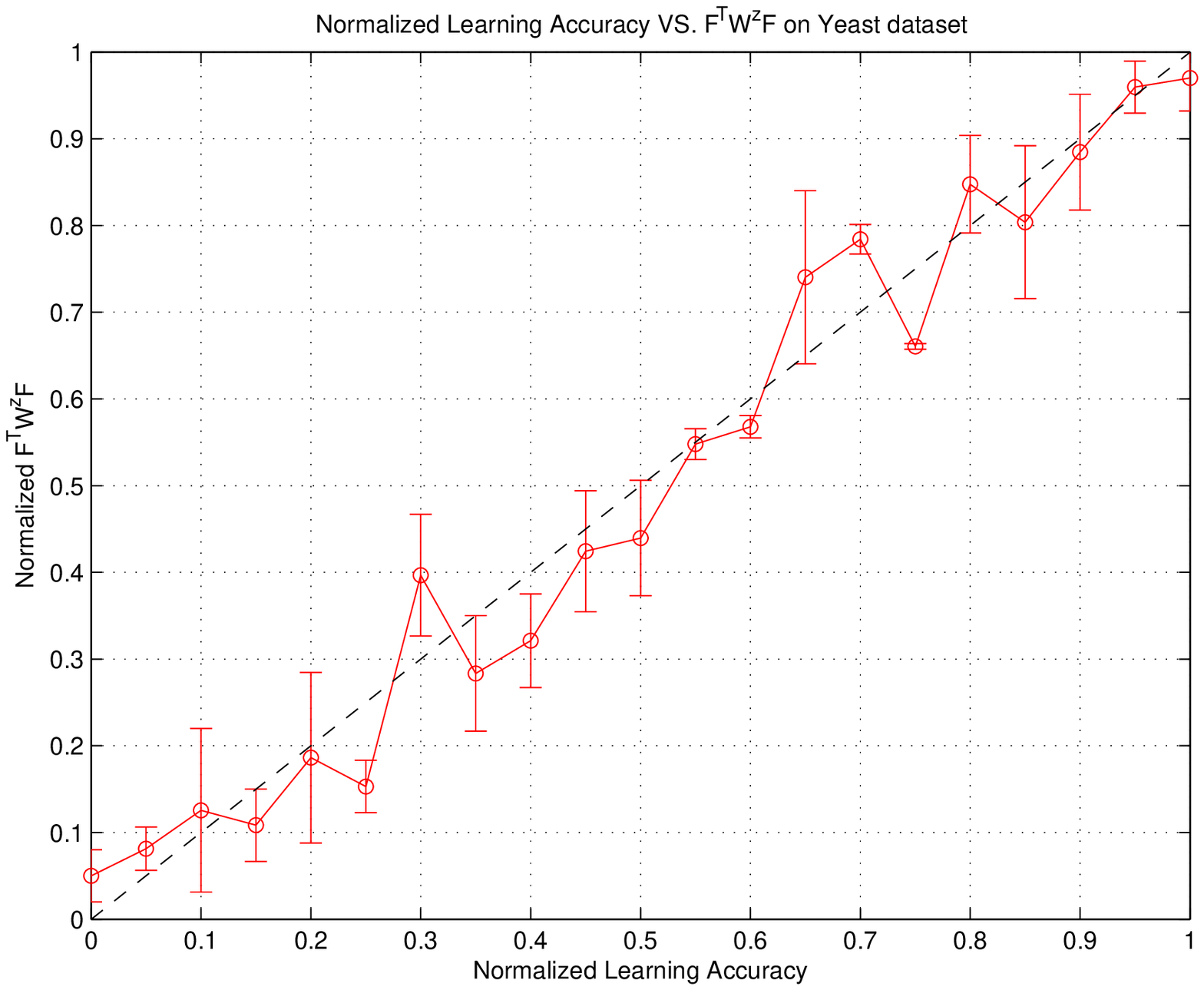}}
\subfigure[Multi-view learning on Caltech-256 dataset]{\label{fig:fwf_2}
\includegraphics*[width=0.46\textwidth]{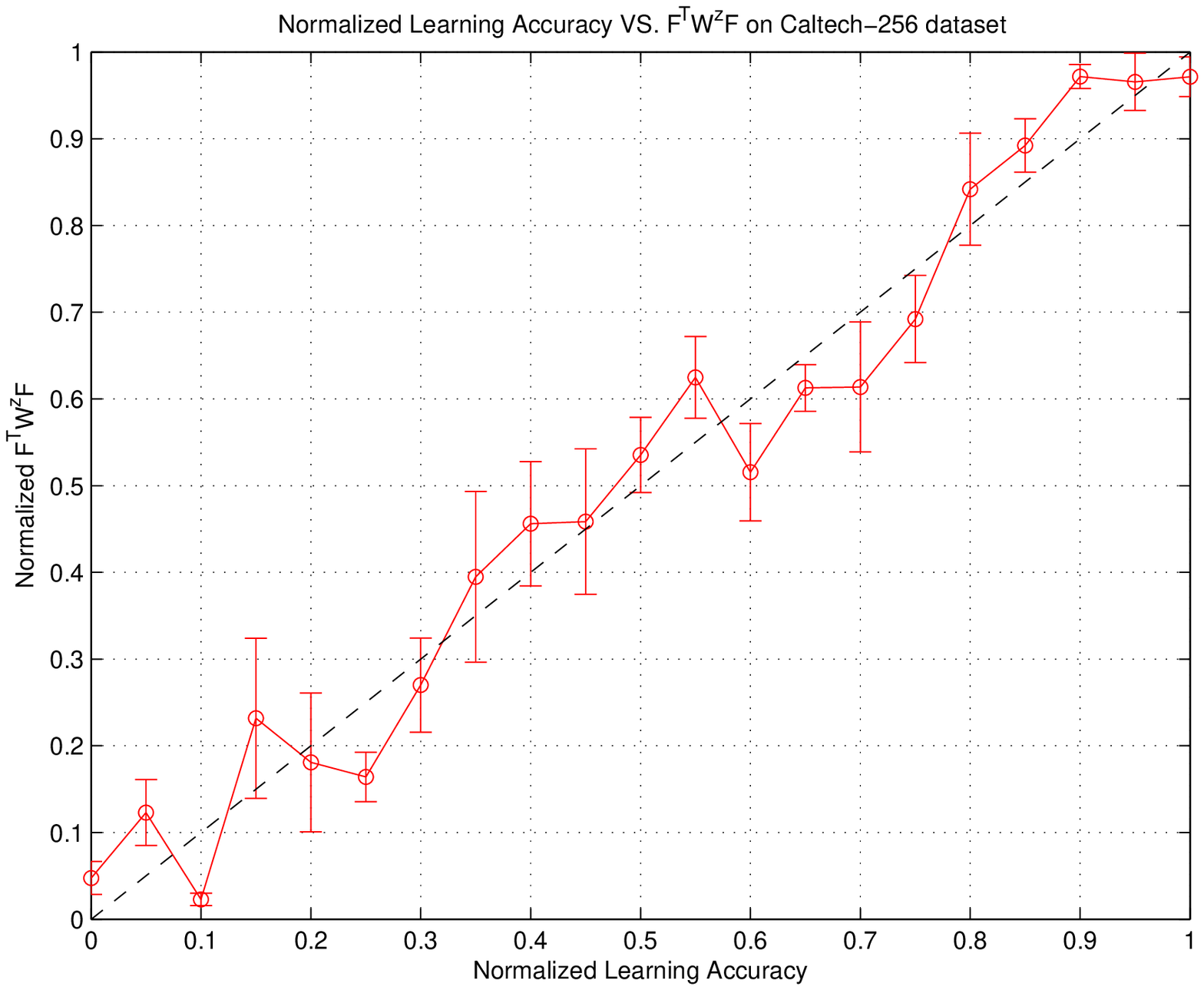}}
\caption{Examples showing the approximately proportional relation between the learning accuracy and the value of $\mathrm{CP} \left(\mathbf{W} \right)$.} \label{fig:fwf}
\end{center}
\end{figure}

\textbf{Empirical Demonstration.} We shall now illustrate the use of the proposed success measure. We apply our approach to two real-world datasets: Yeast gene dataset and Caltech-256 image dataset. Since our algorithms are deterministic, we can obtain a variety of experiments by changing the parameters $\alpha_k$ or $\beta_k$. We collect a set of learning accuracies and the corresponding values of $\mathrm{CP} \left(\mathbf{W} \right)$ under these different parameterizations, and normalize all of them to lie between $0$ and $1$. In this experiment we adopt two tasks (DNA synthesis and transcription) for multi-task learning on Yeast, and two views (visual word and color) for multi-view learning on Caltech-256. Fig.~\ref{fig:fwf_1} shows the learning accuracy v.s. the value of $\mathbf{F}^T \mathbf{W}^z \mathbf{F}$ (sum of off-diagonal values) under multi-task setting on Yeast dataset with respect to different $\beta_k$'s. Fig.~\ref{fig:fwf_2} shows the normalized learning accuracy v.s. the value of $\mathbf{F}^T \mathbf{W}^z \mathbf{F}$ (naturally a single value) under multi-view setting on Caltech-256 dataset with respect to different parameterizations of $\alpha_k$. The results shown in the two figures are quantified and averaged from $100$ random trials (per parameterization). The red solid curve shows the results obtained using our SSDR-MML algorithm, while the black dash line marks the perfect proportional relation. Observing the results, in both settings we see that the value of $\mathbf{F}^T \mathbf{W}^z \mathbf{F}$ is approximately linear to the learning accuracy. Therefore, we conclude that a parameterization with relatively higher value of $\mathbf{F}^T \mathbf{W}^z\mathbf{F}$ will generally lead to a better learning result, and thus $\alpha_k$ or $\beta_k$ should be set to the one which corresponds to the highest value of $\mathbf{F}^T \mathbf{W}^z \mathbf{F}$.

\section{Implementation and Pragmatic Issues} \label{sec:imple}

In this section, we outline issues that we believe make implementation and using of our work easier.

\subsection{Uses for Transfer Learning}
The proposed framework is motivated by the fact that multi-task and multi-view learning may improve the learning performance over the ``single" case by exploiting the complementary knowledge contained in the multiple tasks or views. In multi-task learning, higher learning performance could be obtained if the multiple tasks are highly correlated, while worse performance could happen if the multiple tasks are irrelevant. In multi-view learning, intuitively, the multiple views of feature description are supposed to be neither too far nor too close to each other. There would not be much gain if the multiple feature descriptions are too similar. On the other hand, if the multiple feature descriptions are too different, multi-view learning could even be harmful. Since we do not want to over-constrain the graph to any task or view while still absorbing knowledge from each of them, the proposed approach is in fact a moderate solution that partially fits the graph to each task or view in a weighted fashion.

In some practical cases, we may want to regard one of the multiple tasks or views as the target task or view, and consider others as the source view or task to help it. In that case, we set the $\beta_k$ or $\alpha_k$ of the target task or view to $1$, and the weight of the other tasks or views to lie between $0$ and $1$. Then, the weights of these sources tasks and views can be used to encode the relative ``relatedness" and ``importance" to the main task and view, respectively. For a source task or view, a weight of value $0$ indicates that it is ``irrelevant" or ``contradictory" to the main task or view, while a weight of value $1$ implies that it is ``perfectly correlated" or ``complementary" to the target task or view, respectively. 

\subsection{Parameter Selection}

\begin{table}[!htb] \caption{SSDR-MML Algorithm (progressive)} \label{tab:tab1}
\begin{center}
\begin{tabular}{l}
\hline \hline
\textbf{Input:}\\
\quad \texttt{feature descriptions $\mathbf{x}^k_i$, for $i\in\{1,\cdots,n\}$ and $k\in\{1,\cdots,q\}$} \\
\quad \texttt{prior labels matrix $\mathbf{Y}^k$, for labeled instances and $k\in\{1,\cdots,p\}$} \\
\quad \texttt{regularization parameters $\alpha_k$, $\beta_k$ and $\lambda$.}\\
\\
\textbf{Training Stage:}\\
\qquad \texttt{initialize: count $m=0$, ${(\mathbf{Y}^k)}^0=\mathbf{Y}^k$, $\mathbf{F}^k_l={(\mathbf{Y}^k)}^0$.}\\
\qquad \texttt{do\{}\\
\qquad \qquad \texttt{compute $\mathbf{w}^m_{i\bullet}=\dfrac{\mathbf{1}^T\left(\sum_{k=1}^q \alpha_k C^{\mathbf{x}^k_i} + \sum_{k=1}^p \beta_k C^{\mathbf{f}^k_{i\bullet}} + \lambda I\right)^{-1}}
{\mathbf{1}^T\left(\sum_{k=1}^q \alpha_k C^{\mathbf{x}^k_i} + \sum_{k=1}^p \beta_k C^{\mathbf{f}^k_{i\bullet}} + \lambda I\right)^{-1}\mathbf{1}}$;}\\
\qquad \qquad \texttt{update $\mathbf{D}^{m}$, $d^m_{jj} = \sum_{i=1}^{n}w^{m}_{ij}$;}\\
\qquad \qquad \texttt{update ${(\mathbf{v}^k)}^{m} = \sum_{i=1}^{c^k} \dfrac{{\left(\mathbf{y}^k_{\bullet i}\right)}^{m}\odot \mathbf{D}^{m}\mathbf{1}}{{{\left({\left(\mathbf{y}^k_{\bullet i}\right)}^{m}\right)}^T} \mathbf{D}^{m}\mathbf{1}}$;}\\
\qquad \qquad \texttt{locate $(i^*,j^*)^k = \mathrm{arg\,min}_{i,j} \left(\dfrac{\partial{\mathcal{Q}}}{\partial{\mathbf{F}^k}}\right)_u$;}\\
\qquad \qquad \texttt{set $f^k_{(l^k+i^*)j^*}=1$, and $f^k_{(l^k+i^*)j}=0$ for $j\neq j^*$}; \\
\qquad \qquad \texttt{update ${\left(\mathbf{Y}^k\right)}^{m+1} = \left[\begin{array}{ll}{\left(\mathbf{Y}^k\right)}^{m} \\\mathbf{f}^k_{(l^k+i^*)\bullet}\end{array}\right]$;}\\
\qquad \qquad \texttt{update $\mathbf{F}^k_l={\left(\mathbf{Y}^k\right)}^{m+1}$, and remove the $(l^k+i^*)^{th}$ instance from $\mathbf{F}^k_u$;}\\
\qquad \qquad \texttt{update $m=m+1$, $l^k=l^k+1$;}\\
\qquad \texttt{\}while($\mathbf{F}^k_u != \phi$)}\\
\\
\textbf{Output:}\\
\quad \texttt{weight matrix $\mathbf{W}$, label predictions $\mathbf{Y}^k_u$ for $k\in\{1,\cdots,p\}$.}\\
\hline \hline
\end{tabular}
\end{center}
\end{table} 

The general solution for the proposed algorithm can be implemented as shown in Table \ref{tab:tab1}. There are three parameters in our proposed framework $\alpha_i$ (the weight of view $i$), $\beta_j$ (the weight of task label $j$), and $\lambda$ (the sparsity regularization parameter). The sparsity constraint has two functionings, one is to avoid excessive label propagation, the other one is to simplify the learning model. In practice, we found that the learning performance of our approach is not sensitive to the sparsity controller $\lambda$, which implies that the excessive label propagation can be avoided if $\lambda$ is not too small. Then the selecting of $\lambda$ naturally introduce a trade-off between model simplicity and model-fit. If the simplicity of learning model is preferred, we should choose a larger $\lambda$. On the contrary, if better model-fit is preferred, $\lambda$ should be set to a relatively smaller value.

\begin{figure}[!ht]
\begin{center}
\subfigure[Multi-task learning on Yeast dataset]{\label{fig:para_1}
\includegraphics*[width=0.46\textwidth]{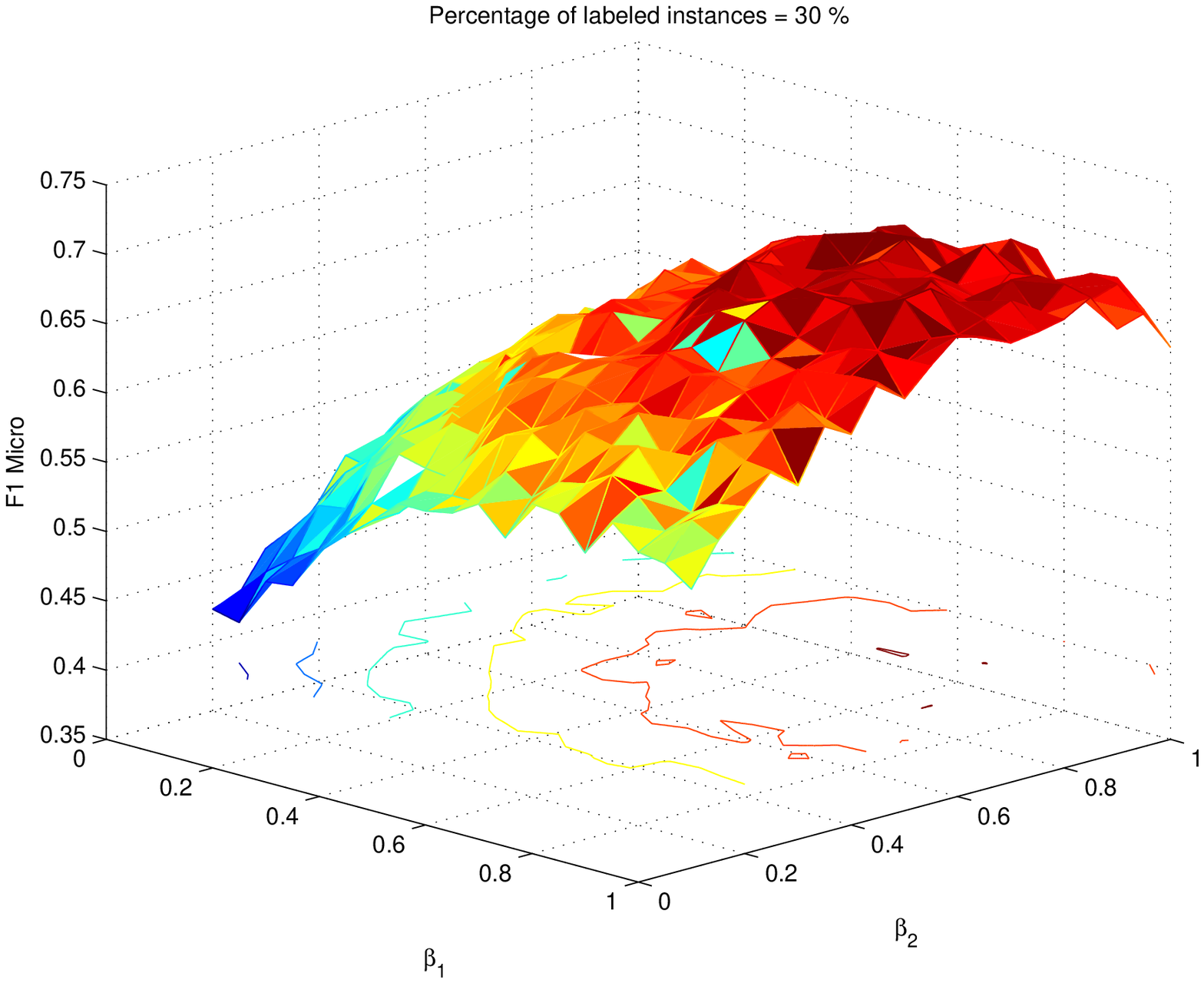}}
\subfigure[Multi-view learning on Caltech-256 dataset]{\label{fig:para_2}
\includegraphics*[width=0.46\textwidth]{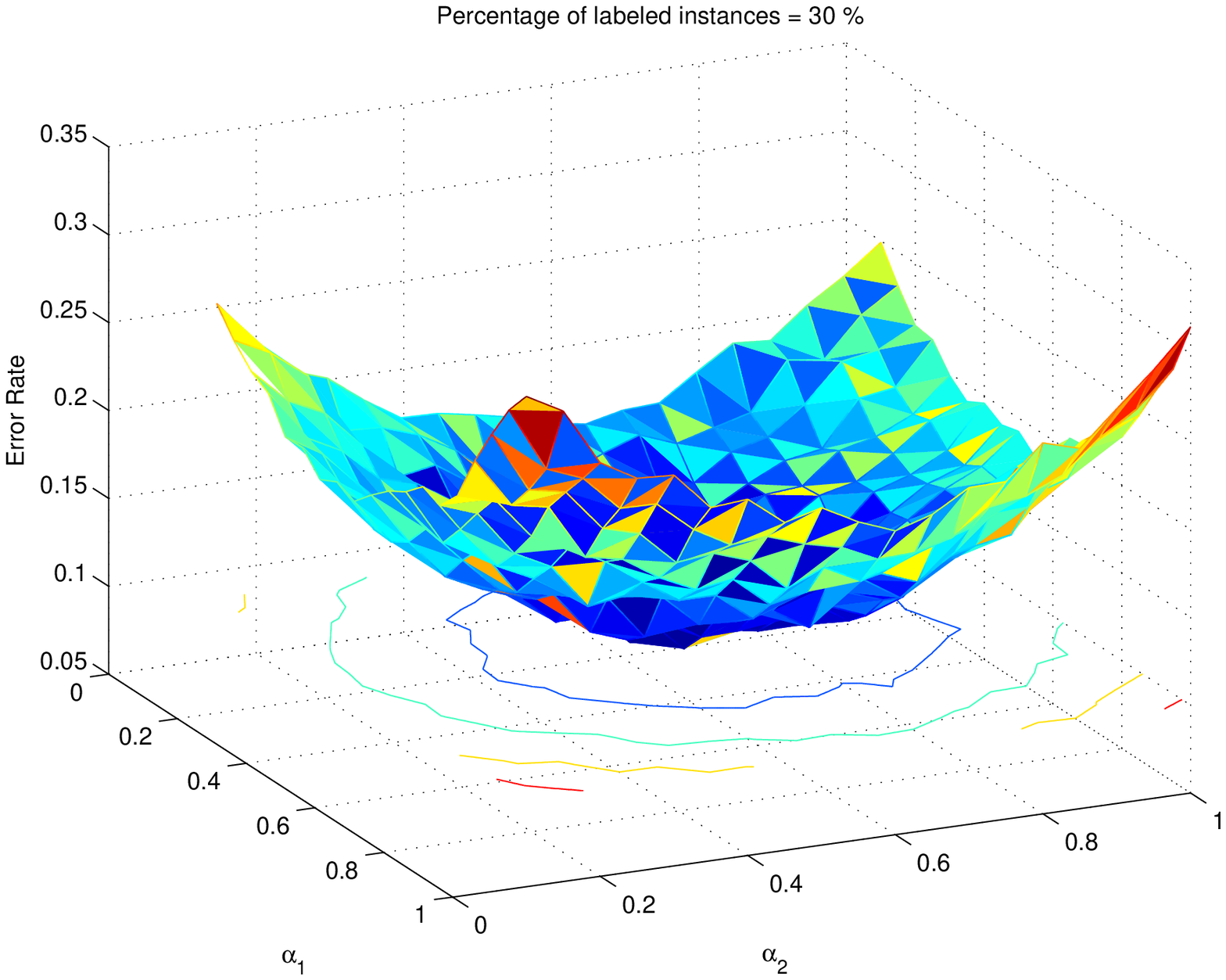}}
\caption{Learning performance of SSDR-MML w.r.t. different parameter settings} \label{fig:para}
\end{center}
\end{figure}

With respect to the the learning performance of our framework to the parameters $\alpha_k$ and $\beta_k$, we wish to show that the performance of the algorithm does not fluctuate greatly with respect to minor changes in the parameter values. We empirically show the stability of the performance of our framework with respect to the parameters ($\alpha_k$ and $\beta_k$) in Fig.~\ref{fig:para}. We first pick two highly related tasks, Cell Growth, Cell Division, DNA synthesis v.s. Transcription, from Yeast dataset (gene dataset with multiple tasks), on which we show the learning performance of our framework in Fig.~\ref{fig:para_1} using $F_1$ Micro score w.r.t. different settings of $\beta_k$ ($0.1\leq \beta_k \leq 1$), where $\beta_1$ and $\beta_2$ are two parameters used to balance the influence of the two tasks. We then choose a subset --- swiss-army-knife v.s. telephone-box --- from Caltech-256 image dataset, and construct two views from the images: (1) visual word histogram using SIFT visual feature where color is ignored by definition; and (2) color histogram. Fig.~\ref{fig:para_2} shows the error rate of our framework on the binary learning task w.r.t. different settings of $\alpha_k$ ($0.1\leq \alpha_k \leq 1$), where $\alpha_1$ and $\alpha_2$ are two parameters used to balance the influence of the two views. In both Fig.~\ref{fig:para_1} and \ref{fig:para_2}, we see that learning performance surface is relatively smooth. 

Though the algorithm's performance is stable, how to set the parameters is still an open question as is the case for most learning algorithms. Two alternatives are to use the knowledge from domain experts to guide or constrain the selection of parameters, e.g. it is well known that the color feature should be more emphasized in the image classification problem ``tomato v.s. washing machine". Alternatively, our previously defined measure for success of learning $\mathrm{CP} (\mathbf{W})$ in section \ref{sec:success} could be used to guide the selection of parameters which is the approach we use in our experimental section.

\subsection{Computational Complexity}
The standard implementation of algorithm shown in Table \ref{tab:tab1} take $\mathcal{O}(mn^3)$ time, where $m$ denotes the number of missing labels in the multiple tasks, and $n$ denotes the total number of instances. Since the fact $m<n$ , the computation complexity of our algorithm is dominated by the matrix inversion operation, where standard methods require $\mathcal{O}(n^3)$ time. To speed up the computation, the weight matrix $\mathbf{W}$ can be recovered by solving linear system of equations as it is in practice much efficient than inversing a matrix. Observing the weight recovering function shown in Eq.~(\ref{eq:rekt5}) or Eq.~(\ref{eq:rekt12}), we see that the denominator of the fraction is a constant which rescales the sum of $i$-th row of $\mathbf{W}$ to $1$. Therefore, in practice, a more efficient way to recover the optimal $\mathbf{w}_{i\bullet}$ is simply to solve the linear system of equations, and then rescale the sum of weights to one. Let $\mathbf{L}_{i}$ denote the mixed local covariance matrix $\left(\sum_{k=1}^q \alpha_k C^{\mathbf{x}^k_i} + \sum_{k=1}^p \beta_k C^{\mathbf{f}^k_{i\bullet}} + \lambda I\right)$. The optimal $\mathbf{w}_{i\bullet}$ can be recovered efficiently by solving the linear system of equations $\mathbf{L}_{i} \mathbf{w}_{i\bullet} = \mathbf{1}$, and then rescale the sum of $\mathbf{w}_{i\bullet}$ to 1. When the local covariance matrices is singular, the linear system of equations can be conditioned by adding a small multiple of the identity matrix
\begin{equation} \label{eq:ssdr-mc13}
\mathbf{L}_{i} \leftarrow \mathbf{L}_{i}+\dfrac{\xi tr(\mathbf{L}_{i})}{k}I
\end{equation}
where $k$ denotes the number of neighbors for each instance, and $\xi$ is a very small value ($\xi\ll0.1$).

\subsection{Extensions for Noisy Prior Labels}
Here we show how a simple extension to our work can allows some incorrect labels to be ignored.
To deal with noisy training data, we provide another solution to update the classifying function $\mathbf{F}^k$, allowing the algorithm to ignore the possibly incorrect given labels. Since we previously assumed that all the initial labels are accurate, the solution of $\mathbf{F}^k$ provided in Eq.~(\ref{eq:rekt7}) or Eq.~(\ref{eq:rekt8}) suffers from a serious problem that there may be considerable noise scattered in the labeled data. A reasonable solution to address this problems is to relax the inference objective function by replacing the constraint on the given labels with an inconsistent penalty term, namely local fitting penalty \cite{lgc} allowing partial neglect of the noisy given labels. We now expand the prior label matrix $\mathbf{Y}^k$ to be a $n\times c^k$ matrix, and fill in the missing label locations with zeros. To relax the inference problem shown in Eq.~(\ref{eq:rekt6}), we add in the inconsistent penalty term and rewrite the cost function as

\begin{eqnarray} \label{eq:rekt17}
\min_{\mathbf{F}^k}\quad \mathcal{Q}(\mathbf{F}^k,\mathbf{Y}^k)=\dfrac{1}{2}tr\left\lbrace \left({\mathbf{F}^k}\right)^T\left(I-\mathbf{W}\right)^T\left(I-\mathbf{W}\right)\mathbf{F}^k + \gamma\left(\mathbf{F}^k-\mathbf{V}^k\mathbf{Y}^k\right)^T\left(\mathbf{F}^k-\mathbf{V}^k\mathbf{Y}^k\right)\right\rbrace
\end{eqnarray} \\
where the coefficient $\gamma > 0$ is a tuning parameter balancing the influence of label reconstruction error and local fitting penalty. If we set $\gamma=\infty$, the cost function will reduce to Eq.~(\ref{eq:rekt6}). The minimization problem is straightforward since the cost function is convex and unconstrained. Then, the update for $\mathbf{F}^k$ shown in Eq.~(\ref{eq:rekt7}) can be rewritten as 

\begin{equation} \label{eq:rekt18}
\dfrac{\partial{\mathcal{Q}}}{\partial{\mathbf{F}^{k}}} = \left(I-\mathbf{W}\right)^T\left(I-\mathbf{W}\right)\mathbf{F}^k+ \gamma\left(\mathbf{F}^k-\mathbf{Y}^k\right)=0 \quad\Longrightarrow\quad \mathbf{F}^k = {\left(\frac{1}{\gamma} \left(I-\mathbf{W}\right)^T \left(I-\mathbf{W}\right) + I\right)}^{-1}\mathbf{V}^k\mathbf{Y}^k
\end{equation}

Accordingly, in this relaxed version of label inference, the progressive update for $\mathbf{Y}^k$ will also change. Since the prior label matrix $\mathbf{Y}^k$ is included in the cost function Eq.~(\ref{eq:rekt17}), the optimization problem is now over both the classifying function $\mathbf{F}^k$ and the prior label matrix $\mathbf{Y}^k$, mathematically $\min_{\mathbf{F}^k, \mathbf{Y}^k} \mathcal{Q}(\mathbf{F}^k, \mathbf{Y}^k)$. Therefore, we replace $\mathbf{F}^k$ in the cost function Eq.~(\ref{eq:rekt17}) with its optimal solution shown in Eq.~(\ref{eq:rekt18}), let $A={\left(\frac{1}{\gamma} \left(I-\mathbf{W}\right)^T \left(I-\mathbf{W}\right) + I\right)}^{-1}$, the optimization problem over $\mathbf{Y}^k$ can be formulated as 

\begin{eqnarray} \label{eq:rekt19}
\mathcal{Q}(\mathbf{Y}^k) &=& \frac{1}{2} tr\left\lbrace {\left(\mathbf{A}\mathbf{V}^k\mathbf{Y}^k\right)}^T {\left(I-\mathbf{W}\right)} ^T \left(I-\mathbf{W}\right) \left(\mathbf{A}\mathbf{V}^k\mathbf{Y}^k\right) + \gamma{\left(\mathbf{A}\mathbf{V}^k\mathbf{Y}^k- \mathbf{V}^k\mathbf{Y}^k\right)}^T \left(\mathbf{A}\mathbf{V}^k\mathbf{Y}^k- \mathbf{V}^k\mathbf{Y}^k\right) \right\rbrace\\ \nonumber
&=& \frac{1}{2} tr\left\lbrace {\left(\mathbf{V}^k\mathbf{Y}^k \right)}^T \left[ \mathbf{A}^T {\left(I-\mathbf{W}\right)}^T \left(I-\mathbf{W}\right) \mathbf{A} + \gamma {\left(\mathbf{A}-I\right)}^T \left(\mathbf{A}-I\right) \right] \left(\mathbf{V}^k\mathbf{Y}^k\right) \right\rbrace
\end{eqnarray}

Let $\mathbf{B} = \mathbf{A}^T {\left(I-\mathbf{W}\right)}^T \left(I-\mathbf{W}\right) \mathbf{A} + \gamma {\left(\mathbf{A}-I\right)}^T \left(\mathbf{A}-I\right)$, we write the gradient of $\mathcal{Q}$ with respect to $\mathbf{V}^k\mathbf{Y}^k$ as show in Eq.~(\ref{eq:rekt20}), which substitutes Eq.~(\ref{eq:rekt8}).

\begin{equation} \label{eq:rekt20}
\frac{\partial{\mathcal{Q}}}{\partial{\mathbf{V}^{k}\mathbf{Y}^{k}}} = \left( \mathbf{B}^T \mathbf{B} \right) \mathbf{V}^{k}\mathbf{Y}^{k}
\end{equation}

Base on $\dfrac{\partial{\mathcal{Q}}}{\partial{\mathbf{Y}^k}} = \dfrac{\partial{\mathbf{V}^k\mathbf{Y}^k}}{\partial{\mathbf{Y}^k}} \dfrac{\partial{\mathcal{Q}}}{\partial{\mathbf{V}^k\mathbf{Y}^k}}$ and basics in algebra, we know that the optimization of $\mathcal{Q}$ with respect to $\mathbf{Y}^k$ is equivalent to the optimization over $\mathbf{V}^k\mathbf{Y}^k$. Consequently, the most confident prediction is located at the position shown in Eq.~(\ref{eq:rekt21}), which substitutes Eq.~(\ref{eq:rekt9}).

\begin{equation} \label{eq:rekt21}
(i^*,j^*)^k = \mathrm{arg\,min}_{i,j} \left(\dfrac{\partial{\mathcal{Q}}}{\partial{\mathbf{V}^k\mathbf{Y}^k}}\right)_u 
\end{equation}

\section{Empirical Study} \label{sec:experiment}
In this section, we empirically evaluate our framework SSDR-MML on several real-world applications under multi-task and multi-view learning settings.

\subsection{Multi-task Learning}

\subsubsection{Experimental Settings}

In this empirical study, we compare the performance of our SSDR-MML approach against four baseline models: (1) \textbf{RankSVM} \cite{ranksvm}, a state-of-the-art supervised multi-label classification algorithm based on ranking the results of support vector machine (SVM); (2) \textbf{PCA+RankSVM}, which performs a principled component analysis (PCA) dimension reduction as a separate step before RankSVM; (3) \textbf{ML-GFHF}, the multi-task version (two-dimensional optimization) of the harmonic function \cite{smse}; (4) \textbf{Regularized MTL} \cite{multitask2}, a regularized multi-task learning, which is based on the minimization of regularization functionals similar to SVM, and assumes that all predicting functions come from a Gaussian distribution. In RankSVM, we choose RBF kernel function ($\sigma$ set to the average of Euclidean distances between all pairs of data points), and fix the penalty coefficient $C=1000$. For ML-GFHF, we construct a $k$-\textsl{NN} ($k=15$) graph similarity via RBF kernel function with length scale $\sigma= \sum_{i=1}^{n} \parallel \mathbf{x}_{i} -\mathbf{x}_{ik} \parallel/n$, where $\mathbf{x}_{ik}$ is the $k$-th nearest neighbor of $\mathbf{x}_{i}$. In our SSDR-MML approach, we set the regularization parameter $\lambda = 1$ and determine the importance of each task by maximizing the learning success measure, $\beta_i = \mathrm{arg} \max_{\beta_i} \mathrm{CP} (\mathbf{W})$. For fairness, the parameters ($\lambda_i$) in Regularized MTL are set to the values that are equivalent to the parameter setting of SSDR-MML. In multi-task learning, we adopt a standard evaluation method for the multiple labels, micro-averaged $F_1$ measure ($F_1$ Micro) \cite{f1micro}.\\

\noindent\textbf{Dataset.} We evaluate the performance of our framework for multi-task learning on three different types of real world datasets. 
\begin{itemize}
\item \textbf{Yeast} \cite{yeast}: consists of $2,417$ gene samples, each of which belongs to one or several of $14$ distinct functional categories, such as transcription, cell communication, protein synthesis, Ionic Homeostasis, and etc. The feature descriptions of \emph{Yeast} dataset are extracted by different sequence recognition algorithms, and each gene sample is represented in a $103$ dimensional space. The tasks in \emph{Yeast} dataset are to predict the localization site of protein, where each sample is associated with $4.24$ labels on average. 

\item \textbf{Scene}: image dataset consists of $2,407$ natural scene images, each of which is represented as a $294$-dimensional feature vector and belongs to one or more (1.07 in average) of 6 categories (beach, sunset, fall foliage, field, mountain, and urban). 

\item \textbf{SIAM TMC 2007}: text dataset for SIAM Text Mining Competition 2007 consisting of $28,596$ text samples, each of which belongs to one or more ($2.21$ in average) of $22$ categories. To simplify the problem, in our experiments we take a randomly selected subset containing $3,000$ samples from the original dataset, then use binary term frequencies and normalize each instance to unit length ($30,438$-dimensional vector). 
\end{itemize}

We chose these three multi-task datasets because that represent a range of situations, most instances in \emph{Yeast} have more than one label while most instance in \emph{Scene} have only one label, and  \emph{SIAM TMC 2007} is a very high dimensional dataset.

\subsubsection{Empirical Result}

\begin{figure}[!ht]
\begin{center}
\subfigure[Yeast]{\label{fig:multitask_yeast}
\includegraphics*[width=0.46\textwidth]{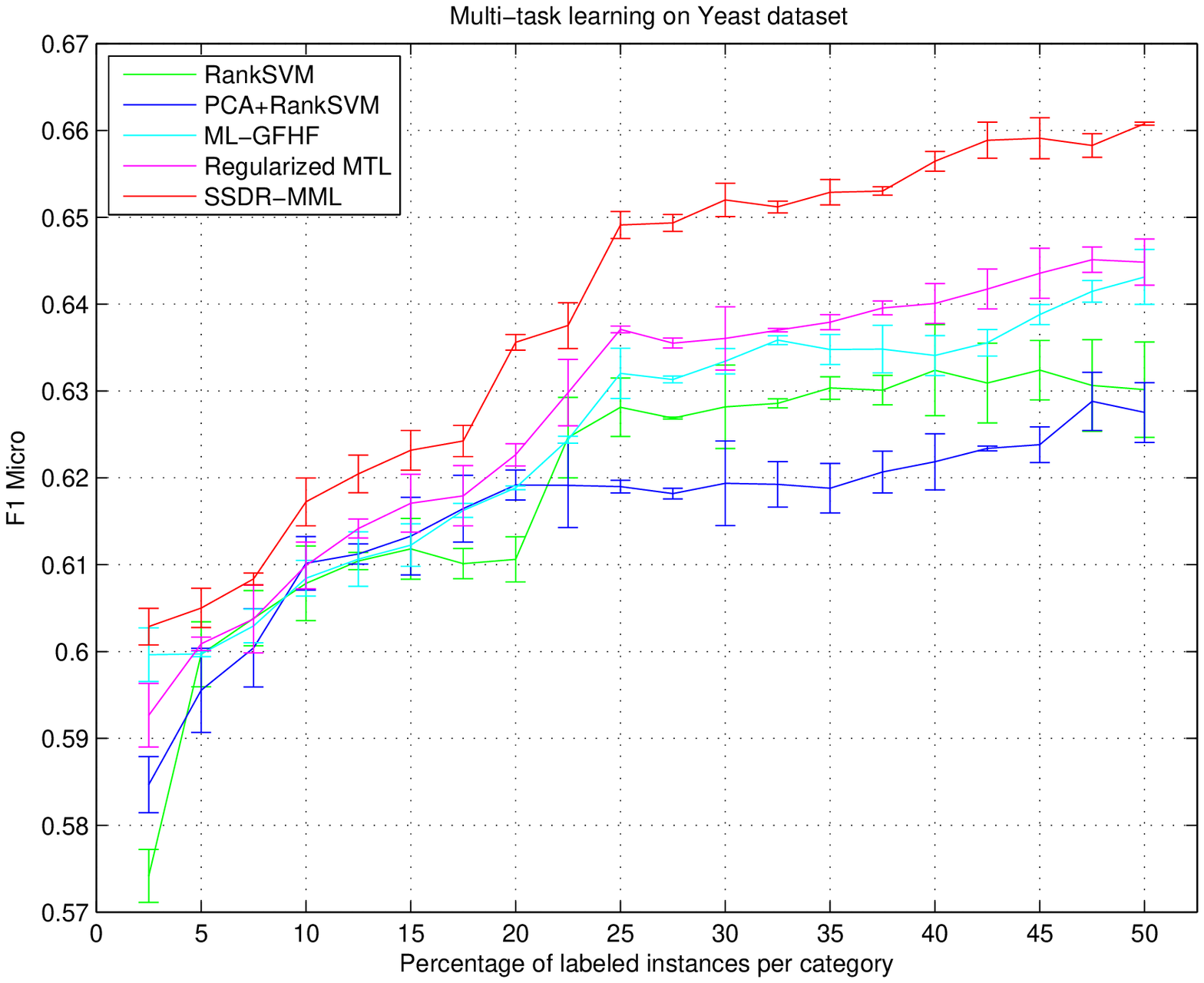}}
\subfigure[Scene]{\label{fig:multitask_scene}
\includegraphics*[width=0.46\textwidth]{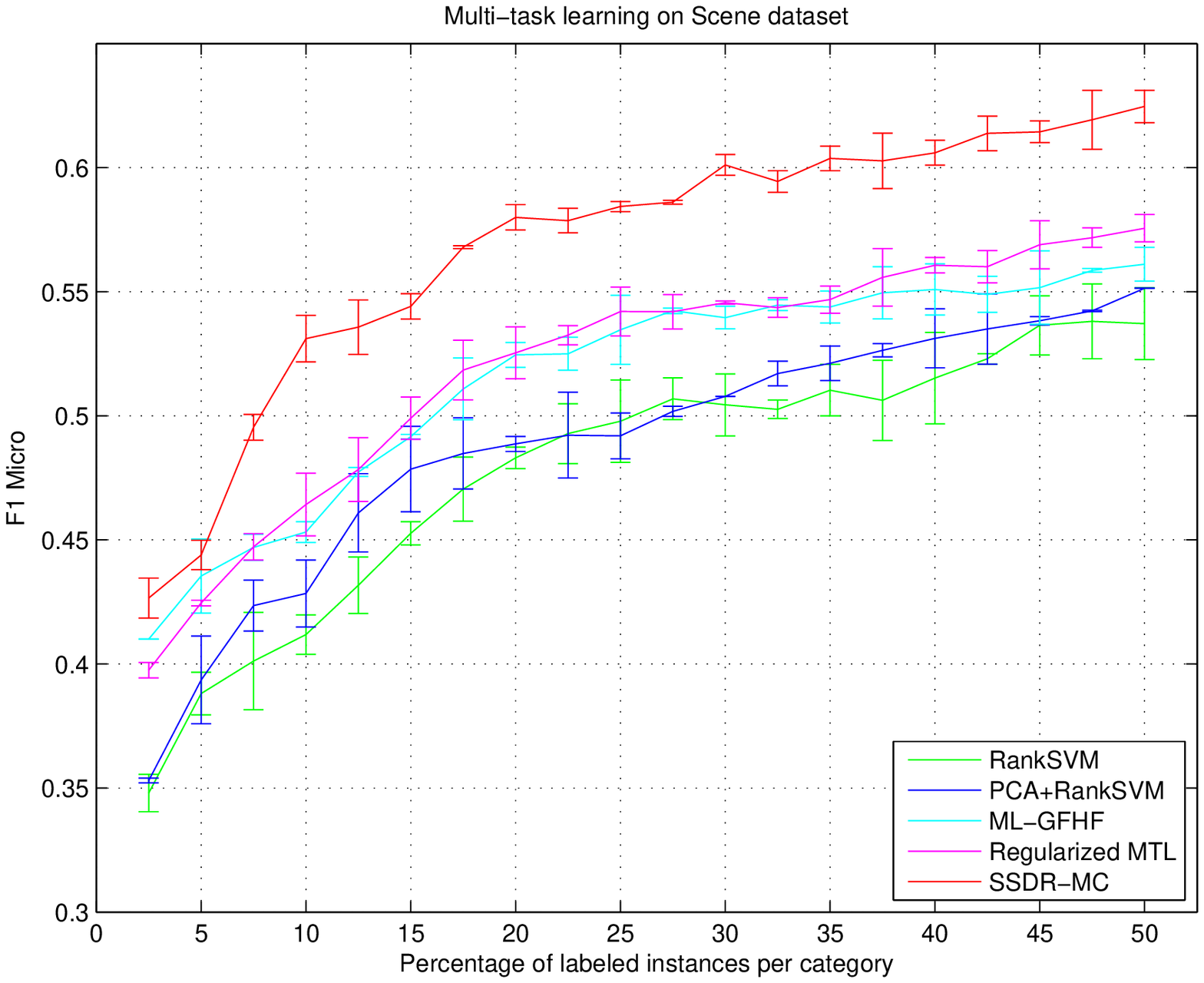}}
\subfigure[SIAM TMC 2007]{\label{fig:multitask_tmc}
\includegraphics*[width=0.46\textwidth]{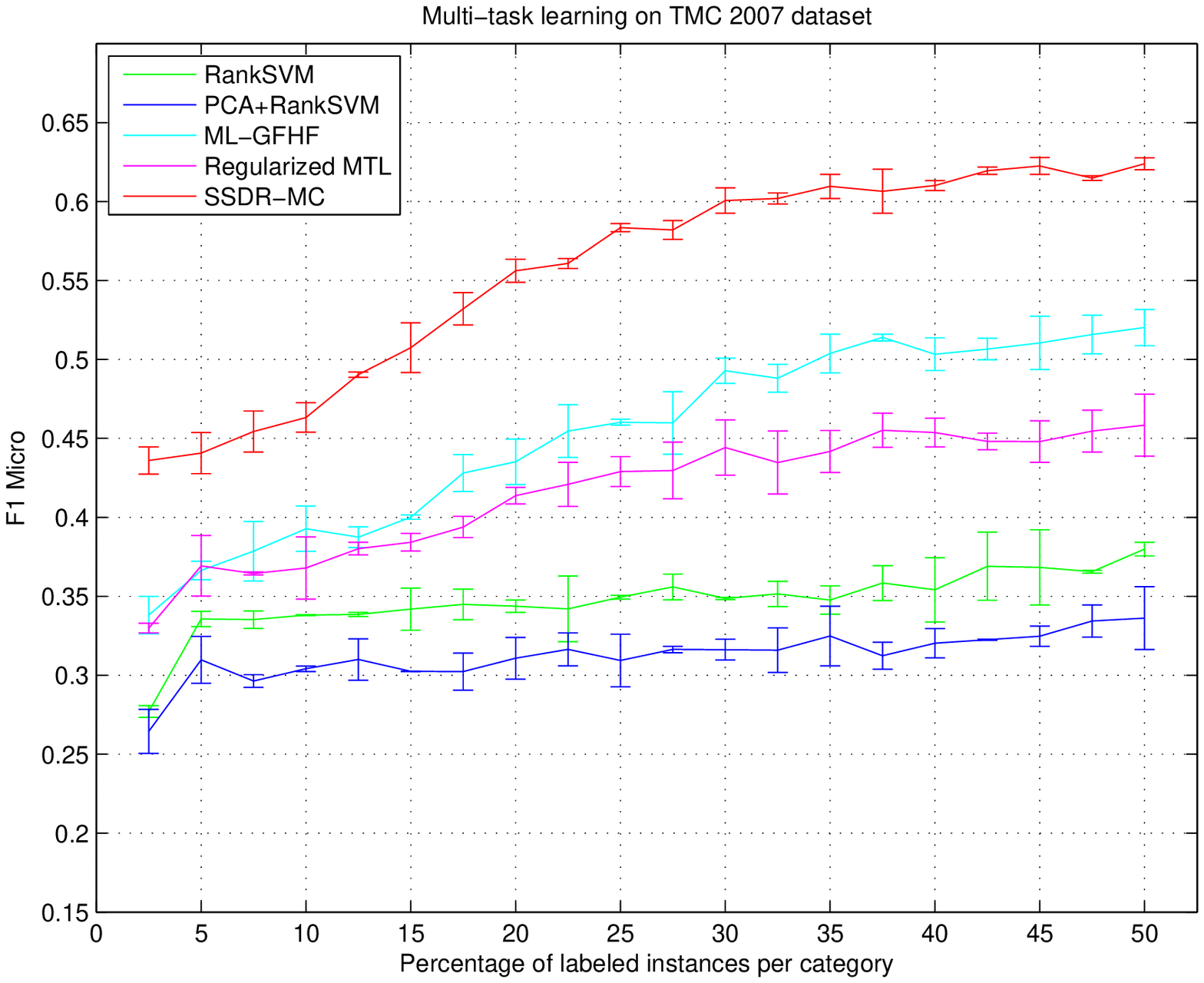}}
\caption{Learning performance measured by $F_1$ Micro score w.r.t. different numbers of labeled data} \label{fig:multitask_result}
\end{center}
\end{figure}

To comprehensively compare our proposed algorithm with the four base methods, we train the algorithms on data with varying numbers of labeled instances. In each trial, for each task we randomly select a portion of instances from the dataset as the training set, while the remaining unlabeled ones are used for testing. The portion of labeled data gradually increases from $2.5\%$ to $50\%$ with a step size of $2.5\%$. In Fig.~\ref{fig:multitask_result}, we report the average $F_1$ Micro scores and standard deviations, all of which are based on the average over 50 trials. For the three real-world datasets that we explored in the experiments, when all experimental results (regardless of step size) are pooled together, the proposed approach SSDR-MML performs statistically significantly better than the competing algorithms at the $98.01\%$ confidence level with both lower error rate and standard deviation. We observe from Fig.~\ref{fig:multitask_tmc} that PCA+RankSVM does not outperform RankSVM on the high-dimensional data SIAM TMC 2007, which indicates that the lack of connection between dimension reduction and learning algorithm would limit the usefulness of apriori dimension reduction techniques. By the promising performance shown in Fig.~\ref{fig:multitask_tmc}, where we iteratively perform learning and spectral embedding on \emph{SIAM TMC 2007} dataset, we demonstrate the effectiveness of connecting dimension reduction and learning when the data is in high-dimensional space.

\subsection{Multi-view Learning}

\subsubsection{Experimental Settings} \label{sec:mvl_setting}

We evaluate our framework for multi-view learning on two applications: (1) image classification on Caltech \cite{caltech} dataset; and (2) a set of UCI benchmarks. We compare the performance of our SSDR-MML approach against three baseline models: (1) \emph{SVM}, standard support vector machine \cite{libsvm} on each view of the data separately. (2) \emph{SKMsmo}, multiple kernel learning \cite{mklsmo} solved by sequential minimal optimization; (3) \emph{Bayesian Co-Training} (BCT) \cite{bct}, a Gaussian process consensus learner. To further understand our framework, we also compare SSDR-MML against two of its variates, i.e. (i) SSDR-MML Simple: SSDR-MML on single-viewed data which is obtained by simply joining the two view features into a single view; and (ii) SSDR-MML nonsparse, our SSDR-MML implementation without sparsity enforcement, which means excessive label propagation could happen. For both SVM and SKMsmo, the kernel is constructed using RBF with length scale $\sigma=\frac{1}{n(n-1)}\sum_{i=1}^{n}\sum_{j\neq i}\parallel\mathbf{x}_{i} - \mathbf{x}_{j} \parallel$, and the penalty coefficient $C0$ is fixed to $100$. For Bayesian Co-Training (BCT), we choose a standard Gaussian process (GP) setting: zeros mean function, Gaussian likelihood function, and isotropic squared exponential covariance function without latent scale. The hyperparameters in GP models can be adapted to the training data, thus the predictions of Bayesian Co-Training can be readily obtained by consenting the predictions of GP from different views using the learned noise variances. To avoid overfitting, the number of function evaluations in gradient-based optimization is limited to a maximum of $100$. In our SSDR-MML approach, we set the regularization parameter $\lambda = 1$, and the importance of each view is decided by maximizing the learning success measure, $\alpha_i = \mathrm{arg}\max_{\alpha_i} \mathrm{CP}(\mathbf{W})$.

\subsubsection{Color-aided SIFT Matching} \label{sec:caltech}
\emph{Caltech-256} image dataset \cite{caltech} consists of $30,608$ images in $256$ categories, where each category contains around $100$ images on average. We conduct the experiments on ten binary tasks that are randomly selected from Caltech-256, as summarized in Table~\ref{tab:caltech_tasks}. Since images are difficult to describe, there are several standard methods to extract features from an image, such as edge direction, and color or visual word histogram. However, there is no straightforward answer to tell what kind of features can outperform others, as the performance of each type of features highly depends on the specific applications. Thus, it could be desirable if we can make use of multiple image descriptors to help the learner. In the experiment, we exploit two image descriptions for each image: (1) color histogram, a representation of the distribution of colors in an image; and (2) visual word histogram, which is based on the successful image descriptor SIFT \cite{sift}. Although SIFT can accurately detect and describe interesting points in an image, it suffers from the limitation that information carried by color is ignored. Consequently, higher learning accuracy can be expected if we are able to perform learning on both SIFT and color simultaneously. In the preprocessing of the data, we construct the color histogram ($80$ bins in HSV color space) by counting the number of pixels that have colors in each of a fixed list of color ranges that span the color space. To produce the visual word histogram, we first build a visual vocabulary ($800$ visual words) by clustering all the SIFT descriptions collected from the image dataset, and then the histograms can be obtained by mapping images to the visual codebook, which is often called bag-of-word.

\begin{table}[!ht] \caption{Ten binary tasks selected from Caltech-256} \label{tab:caltech_tasks}
\begin{center}
\begin{tabular}{|l||c||c|}
  \hline
Task No. & Task Name & Data Size \\
  \hline
  \hline
Task-1 & binocular vs killer-whale & $216$ vs $91$  \\ \hline
Task-2 & breadmaker vs telephone-box & $141$ vs $84$ \\ \hline
Task-3 & eyeglasses vs skyscraper & $82$ vs $95$ \\ \hline
Task-4 & fireworks vs self-propelled lawnmower & $100$ vs $120$ \\ \hline
Task-5 & mars vs saturn & $155$ vs $92$ \\ \hline
Task-6 & pci-card vs sunflower & $105$ vs $80$ \\ \hline
Task-7 & swiss-army-knife vs telephone-box & $109$ vs $84$ \\ \hline
Task-8 & tennis-ball vs zebra & $98$ vs $96$ \\ \hline
Task-9 & tomato vs washing machine & $103$ vs $84$ \\ \hline
Task-10 & video-projector vs watermelon & $97$ vs $93$ \\
  \hline
\end{tabular}
\end{center}
\end{table}

We compare the performance of our approach against the three baseline techniques on the ten predefined binary tasks. In each trial, we randomly select $10\%$ of the images as the training set, and the rest of the images are used as the test set. The experiment are repeated $50$ times, and the mean error rates and standard deviations are reported in Table~\ref{tab:caltech_results}. It can be seen that in general the proposed SSDR-MML approach outperforms all the competing techniques. Moreover, it performs statistically significantly better than baseline models at the $94.60\%$ confidence level with both lower mean error rate and standard deviation. In addition, as we can observe from the result, the multi-view learning algorithms generally perform better than any of the single-view learners. This confirms the motivation of multi-view transfer that using the knowledge carried by all available feature descriptions to improve the performance of the learner. As expected, SIFT is more discriminative in some tasks, while color is more discriminative in the other tasks. Sometimes, the performances of SIFT and color feature are dramatically different. For example, in the task named ``tennis-ball vs zebra" color feature is much reliable than SIFT, while in the task named ``video-projector vs watermelon" SIFT performs significantly better than color. This demonstrate the necessity of multi-view transfer, as it is generally difficult for the users to tell which kind of features can outperform others. By comparing the performance of SSDR-MML with its two variates, we see that (1) simply placing features from multiple views into a single view does not work well due to the increased dimension and normalization issues and (2) Sparsity need to be enforced in semi-supervised settings since excessive label propagation is generally detrimental and unreliable.

\begin{table*}[!ht]
\begin{footnotesize}
\begin{center}
\begin{tabular}{|l||c|c|c|c|c|c|c|}
  \hline
 Task No. & SVM-SIFT & SVM-color & SKMsmo & BCT & SSDR-MML Simple & SSDR-MML nonsparse & SSDR-MML \\
  \hline
  \hline
Task-1 & $14.05\pm 3.75$ & $11.45\pm 3.16$ & $11.26\pm 3.20$ & $11.63\pm 3.32$ & $11.73\pm 3.63$ & $15.09\pm 5.37$ & $\textbf{ 9.14}\pm 2.55$\\ \hline
Task-2 & $13.77\pm 3.97$ & $15.44\pm 3.86$ & $15.20\pm 3.89$ & $14.78\pm 3.88$ & $13.95\pm 3.91$ & $16.31\pm 4.07$ &$\textbf{12.08}\pm 3.15$\\ \hline
Task-3 & $23.80\pm 8.00$ & $19.81\pm 3.76$ & $19.84\pm 3.58$ & $19.85\pm 3.28$ & $21.27\pm 5.21$ & $23.83\pm 8.22$ &$        19.17 \pm 3.26$\\ \hline
Task-4 & $ 7.11\pm 2.47$ & $ 5.55\pm 1.91$ & $ 5.52\pm 1.89$ & $ 4.88\pm 1.93$ & $ 5.21\pm 2.01$ & $ 5.39\pm 3.84$ &$         3.99 \pm 2.14$\\ \hline
Task-5 & $24.54\pm 6.02$ & $29.72\pm 4.96$ & $25.17\pm 3.13$ & $24.26\pm 4.20$ & $27.58\pm 4.79$ & $31.75\pm 6.45$ &$\textbf{21.80}\pm 3.13$\\ \hline
Task-6 & $11.77\pm 4.77$ & $15.04\pm 3.98$ & $14.55\pm 3.93$ & $13.43\pm 4.48$ & $13.92\pm 4.38$ & $16.08\pm 5.03$ &$\textbf{11.33}\pm 3.93$\\ \hline
Task-7 & $16.39\pm 2.72$ & $13.45\pm 3.48$ & $13.40\pm 3.50$ & $14.36\pm 4.45$ & $15.14\pm 3.36$ & $16.77\pm 4.89$ &$\textbf{13.48}\pm 3.03$\\ \hline
Task-8 & $ 8.91\pm 1.81$ & $23.50\pm 5.22$ & $22.56\pm 5.06$ & $21.29\pm 4.98$ & $20.25\pm 4.07$ & $14.43\pm 5.21$ &$\textbf{10.10}\pm 2.13$\\ \hline
Task-9 & $25.27\pm 6.70$ & $17.80\pm 4.54$ & $17.76\pm 4.72$ & $17.42\pm 4.74$ & $19.61\pm 5.13$ & $17.98\pm 6.36$ &$\textbf{14.56}\pm 4.66$\\ \hline
Task-10 &$18.95\pm 6.68$ & $12.94\pm 4.38$ & $12.57\pm 4.17$ & $15.24\pm 6.14$ & $16.43\pm 5.02$ & $16.27\pm 6.18$ &$\textbf{12.14}\pm 4.90$\\ \hline
\end{tabular}
\caption{Mean error rates and standard deviations (in percentage) of the ten binary tasks from Caltech-256. The statistically significant (at $94.60\%$ confidence level) best result is shown in bold.} \label{tab:caltech_results} 
\end{center}
\end{footnotesize}
\end{table*}

\subsubsection{UCI Benchmarks}

The second evaluation of multi-view learning is carried out on a set of \emph{UCI benchmarks} \cite{uci}, namely (1)\emph{Adult} (subset): extracted from the census bureau database; (2) \emph{Diabetes}: contains the distribution for $70$ sets of data recorded on diabetes patients; (3) \emph{Ionosphere}: radar data was collected by a system that consists of a phased array of $16$ high-frequency antennas with a total transmitted power on the order of $6.4$ kilowatts; (4) \emph{Liver Disorders}: attributes are collected from blood tests which are thought to be sensitive to liver disorders; (5) \emph{Sonar}: contains signals obtained by bouncing sonar signals off a metal cylinder at various angles and under various conditions.; and (6) \emph{SPECT Heart}: describes diagnosing of cardiac Single Proton Emission Computed Tomography (SPECT) images. We chose these six datasets since they have been widely used in the evaluations of various learning algorithms. To create two views for each of the datasets, we equally divide the features of each dataset into two disjoint subsets such that they are related but different, and thus each subset can be considered as one view. The details of the six UCI benchmarks are summarized in Table~\ref{tab:uci_tasks}.

\begin{table}[!ht] \caption{Statistics of UCI benchmarks} \label{tab:uci_tasks}
\begin{center}
\begin{tabular}{|c||c|c|c|}
  \hline
\scriptsize Dataset Name &\scriptsize Instance No. &\scriptsize View $1$ Feature No. &\scriptsize View $2$ Feature No.\\ 
  \hline
  \hline
\scriptsize Adult (subset) & $1,605$ & $60$ & $59$\\
  \hline
\scriptsize Diabetes & $768$ & $4$ & $4$\\  
  \hline
\scriptsize Ionosphere & $351$ & $17$  & $17$ \\  
  \hline
\scriptsize Liver Disorders & $345$ & $3$  & $3$ \\  
  \hline
\scriptsize Sonar & $208$ & $30$  & $30$ \\  
  \hline
\scriptsize SPECT Heart & $270$ & $7$  & $6$ \\  
  \hline  
\end{tabular}
\end{center}
\end{table}

We follow the previously stated methodology, where $10\%$ of the samples are used for training and the remaining $90\%$ for testing, to evaluate the performance of our SSDR-MML framework on multi-view setting. The resulting mean error rates and standard deviations on the six UCI benchmarks, which are based on $50$ random trials, are reported in Table~\ref{tab:uci_results}. It can be observed that the multiple view learning techniques generally outperform the single view classifiers, which substantiates the benefits of learning with multiple views. Among all techniques evaluated in the experiment, our SSDR-MML approach performs statistically significantly better than all competitors at the $92.60\%$ confidence level with both lower misclassification rates and standard deviations. The performance of the proposed framework on multi-view learning tasks not only demonstrates the effectiveness of our approach, but also validates the advantage of simultaneous learning of multiple feature descriptions. On the two variates of our approach, we also see that excessive label propagation and simply joining the multiple views into one could be harmful to the learning performance, which confirms the conclusion made in Section~\ref{sec:caltech}.

\begin{table*}[!ht]
\begin{scriptsize}
\begin{center}
\begin{tabular}{|c||c|c|c|c|c|c|c|}
  \hline
 Datasets & SVM-SIFT & SVM-color & SKMsmo & BCT & SSDR-MML Simple & SSDR-MML nonsparse & SSDR-MML \\
  \hline
  \hline
Adult(subset) &   $22.37\pm 1.50$ & $24.23\pm 1.55$ & $21.24\pm 1.32$ & $19.98\pm 1.68$ & $21.98\pm 1.61$ & $23.38\pm 2.73$ & $        19.40 \pm 1.16$\\ \hline
Diabetes &        $35.22\pm 2.04$ & $35.84\pm 3.14$ & $32.71\pm 2.88$ & $29.27\pm 2.54$ & $31.57\pm 2.93$ & $36.62\pm 3.04$ & $\textbf{27.19}\pm 2.35$\\ \hline
Ionosphere &      $10.70\pm 4.73$ & $19.57\pm 5.33$ & $12.55\pm 4.86$ & $17.41\pm 4.75$ & $13.76\pm 4.97$ & $15.29\pm 5.21$ & $        11.48 \pm 4.51$\\ \hline
Liver Disorders & $45.87\pm 3.43$ & $45.05\pm 3.83$ & $45.47\pm 3.66$ & $58.20\pm 0.82$ & $46.01\pm 3.41$ & $50.32\pm 4.39$ & $\textbf{40.99}\pm 2.75$\\ \hline
Sonar &           $32.91\pm 4.97$ & $34.91\pm 3.64$ & $33.18\pm 4.90$ & $34.17\pm 6.84$ & $33.35\pm 4.26$ & $34.58\pm 5.97$ & $\textbf{31.40}\pm 4.19$\\ \hline
SPECT Heart &     $36.47\pm 6.66$ & $26.77\pm 4.75$ & $29.07\pm 4.22$ & $31.42\pm 4.28$ & $29.69\pm 4.52$ & $29.49\pm 5.18$ & $\textbf{25.83}\pm 4.18$\\ \hline
\end{tabular}
\caption{Mean error rates and standard deviations (in percentage) on UCI benchmarks. The statistically significant (at $92.60\%$ confidence level) best result is shown in bold.} \label{tab:uci_results}
\end{center}
\end{scriptsize}
\end{table*}

Based on the promising experiment results, we conclude that the proposed SSDR-MML framework is advantageous when (1) there are multiple feature descriptions available for each instance that are neither too different nor too similar; or (2) the multiple learning tasks are highly related; or (3) the data is sparsely labeled and in high dimensional space.

\section{Related Work} \label{sec:rework}

Our work is related to four machine learning and data mining topics: \emph{multi-task learning}, \emph{multi-view learning}, \emph{multi-task dimension reduction}, and \emph{semi-supervised learning}. Here we briefly review some related works in the four areas.

\textbf{Multi-task Learning.}
Multi-task learning is motivated by the fact that a real world object naturally involves multiple related attributes, and thereby investigating them together could improve the total learning performance. MTL learns a problem together with other related problems at the same time, that allows the learner to use the commonality among the tasks. The hope is that by learning multiple tasks simultaneously one can improve performance over the ``\emph{no transfer}" case. MTL has been studied from many different perspectives, such as neural networks among similar tasks \cite{multitask1}, kernel methods and regularization networks \cite{multitask2}, modeling task relatedness \cite{taskrelatedness}, and probabilistic models in Gaussian process \cite{gpmt,mtgp} and Dirichlet process \cite{mldirichlet}. Although MTL techniques have been successfully applied to many real world applications, their usefulness are significantly weakened by the underlying relatedness assumption, while in practice some tasks are indeed unrelated and could induce destructive information to the learner. In this work, we propose a measure to quantify the success of learning, as to benefit from related tasks and reject the combining of unrelated (detrimental) tasks. 

\textbf{Multi-view Learning.}
Practical learning problems often involves datasets that are naturally comprised of multiple views. MVL learns a problem together with multiple feature spaces at the same time, that allows the learner to perceive different perspectives of the data in order to enrich the total information about the learning task at hand. \cite{bound} has shown that the error rate on unseen test samples can be upper bounded by the disagreement between the classification-decisions obtained from the independent characterizations of the data. Therefore, as a branch of MVL, co-training \cite{cotrain,bct} aims at minimizing the misclassification rate indirectly by reducing the rate of disagreement among the base classifiers. Multiple kernel learning was recently introduced by \cite{ksdp}, where the kernels are linearly combined in a SVM framework and the optimization is performed as an semidefinite program or quadratically constrained quadratic program. \cite{mklsmo} reformed the problems as a block $l_1$ formulation in conjunction with Moreau-Yosida regularization, so that efficient gradient based optimization could be performed using sequential minimal optimization techniques while still generating a sparse solution. \cite{gemkl} preserved the block $l_1$ regularization but reformulated the problem as a semi-infinite linear problem, which can be efficiently solved by recycling the standard SVM implementations and made it applicable to large scale problems. Although the successes of MVL, many existing approaches suffer from their own limitations: conditional independent assumption is important for co-training both theoretically and empirically \cite{cowork} but it rarely holds in real-world applications; multi-kernel machines are limited to combining multiple kernels in linear manners, and such linear scheme sometimes induces poor data representations. In contrast, our SSDR-MML framework does not require the conditional independence assumption, and the multiple views are fused using nonlinear method (matrix inverse).

\textbf{Multi-task Dimension Reduction.}
Various dimension reduction methods have been proposed to simplify learning problems, which generally fall into three categories: unsupervised, supervised, and semi-supervised. In contrast to traditional classification tasks where classes are mutually exclusive, the classes in multi-task/label learning are actually overlapped and correlated. Thus, two specialized multi-task dimension reduction algorithms have been proposed in \cite{mddm} and \cite{mlsi}, both of which try to capture the correlations between multiple tasks. However, the usefulness of such methods is dramatically limited by requiring complete label knowledge, which is very expensive to obtain and even impossible for those extremely large dataset, e.g. web images annotation. In order to utilize unlabeled data, there are many semi-supervised multi-label learning algorithms have been proposed \cite{smse} \cite{hslmc}, which solve learning problem by optimizing the objective function over graph or hypergraph. However, the performance of such approach is weakened by the lack of the concatenation of dimension reduction and learning algorithm. To the best of our knowledge, \cite{ldrmc} is the first attempt to connect dimension reduction and multi-task/label learning, but it suffers from the inability of utilizing unlabeled data.

\textbf{Semi-supervised learning.}
The study of semi-supervised learning is motivated by the fact that while labeled data are often scarce and expensive to obtain, unlabeled data are usually abundant and easy to obtain. It mainly aims to address the problem where the labeled data are too few to build a good classifier by using the large amount of unlabeled data. Among various semi-supervised learning approaches, graph transduction has attracted an increasing amount of interest: \cite{gfhf} introduces an approach based on a random field model defined on a weighted graph over both the unlabeled and labeled data; \cite{lgc} proposes a classifying function which is sufficiently smooth with respect to the intrinsic structure collectively revealed by known labeled and unlabeled points. \cite{weibuyue} extend the formulation by inducing spectral kernel learning to semi-supervised learning, as to allow the graph adaptation during the label diffusion process. Another interesting direction for semi-supervised learning is proposed in \cite{ssgp}, where the learning with unlabeled data is performed in the context of Gaussian process. The encouraging results of many proposed algorithms demonstrate the effectiveness of using unlabeled data. A comprehensive survey on semi-supervised learning can be found in \cite{semisurvey}.

\section{Conclusion} \label{sec:conclu}
As applications in machine learning and data mining move towards demanding domains, they must beyond the restrictions of complete supervision, single-label, single-view and low-dimensional data. In this paper we present a joint learning framework based on reconstruction error, which is designed to handle both multi-task and multi-view learning settings. It can be viewed as simultaneously solving for two sets of unknowns: filling in missing labels and identifying projection vectors that makes points with similar labels close together and points with different labels far apart. As to improve the learning performance, the underlying objective is to enable the learner to absorb knowledge from multiple tasks and views by partially fitting the graph to each of them. Empirically, our proposed approach was shown to give more reliable results on real world applications with both lower error rate and standard deviation compared to the baseline models. Perhaps the most useful part of our approach is that since the mechanism for combining knowledge is explicit it also offers the ability to measure the success of learning.

\bibliographystyle{latex8}
\bibliography{SSDR-MML.bib}

\end{document}